\documentclass[10pt,twocolumn,letterpaper]{article}

\usepackage{cvpr}              

\usepackage[most]{tcolorbox} 
\usepackage[table]{xcolor}
\usepackage{tikz}
\usepackage{multirow}
\usetikzlibrary{arrows.meta,positioning,shapes.geometric,fit,backgrounds,decorations.pathreplacing}
\usepackage{amsmath,amssymb,amsthm}
\usepackage{subcaption}
\usepackage{wrapfig}
\usepackage{capt-of} 
\usepackage{booktabs,siunitx}
\definecolor{lightblue}{RGB}{227, 244, 254}

\newcommand{\method}{\textsc{Care}}

\newcommand{\acc}{\mathrm{acc}}
\newcommand{\fmt}{\mathrm{fmt}}
\usepackage{pifont}
\newcommand{\cmark}{\ding{51}} 
\newcommand{\xmark}{\ding{55}} 

\definecolor{cvprblue}{rgb}{0.21,0.49,0.74}
\usepackage[pagebackref,breaklinks,colorlinks,allcolors=cvprblue]{hyperref}
\usepackage{amsmath,amssymb,mathtools}
\usepackage{algorithm,algpseudocode}
\usepackage{booktabs}
\usepackage[table]{xcolor}

\newcommand{\github}{\raisebox{-0.5pt}{\includegraphics[height=1.05em]{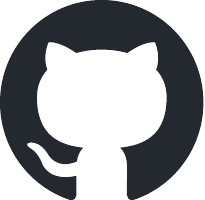}}\xspace}

\def\paperID{} 
\def\confName{CVPR}
\def\confYear{2026}

\title{\method~ What Fails: Contrastive Anchored‑REflection for Verifiable Multimodal Reasoning}

\author{Yongxin Wang$^{1}$, Zhicheng Yang$^{2}$, Meng Cao$^{1}$, Mingfei Han$^{1}$, Haokun Lin$^{1}$, 
\\
Yingying Zhu$^{3}$, Xiaojun Chang$^{1}$, Xiaodan Liang$^{1}$\\
\\
$^{1}$ Mohamed bin Zayed University of Artificial Intelligence \\
$^{2}$ The Hong Kong University of Science and Technology (Guangzhou) 
$^{3}$ Transsion \\
}

\author{
Yongxin Wang$^{1}$ \quad
Zhicheng Yang$^{2}$ \quad
Meng Cao$^{1}$ \quad
Mingfei Han$^{1}$ \quad
Haokun Lin$^{1}$ \quad \\
Yingying Zhu$^{3}$ \quad
Xiaojun Chang$^{1}$ \quad
Xiaodan Liang$^{1}$ \\
\\
$^{1}$Mohamed bin Zayed University of Artificial Intelligence \\
$^{2}$The Hong Kong University of Science and Technology (Guangzhou) \quad
$^{3}$Transsion
}

\begin{document}

\twocolumn[{
\renewcommand\twocolumn[1][]{#1}%
\maketitle

\begin{center}

    \github \url{https://github.com/yongxinwang-ai/CARE}
\end{center}

\begin{center}
  \includegraphics[width=\textwidth]{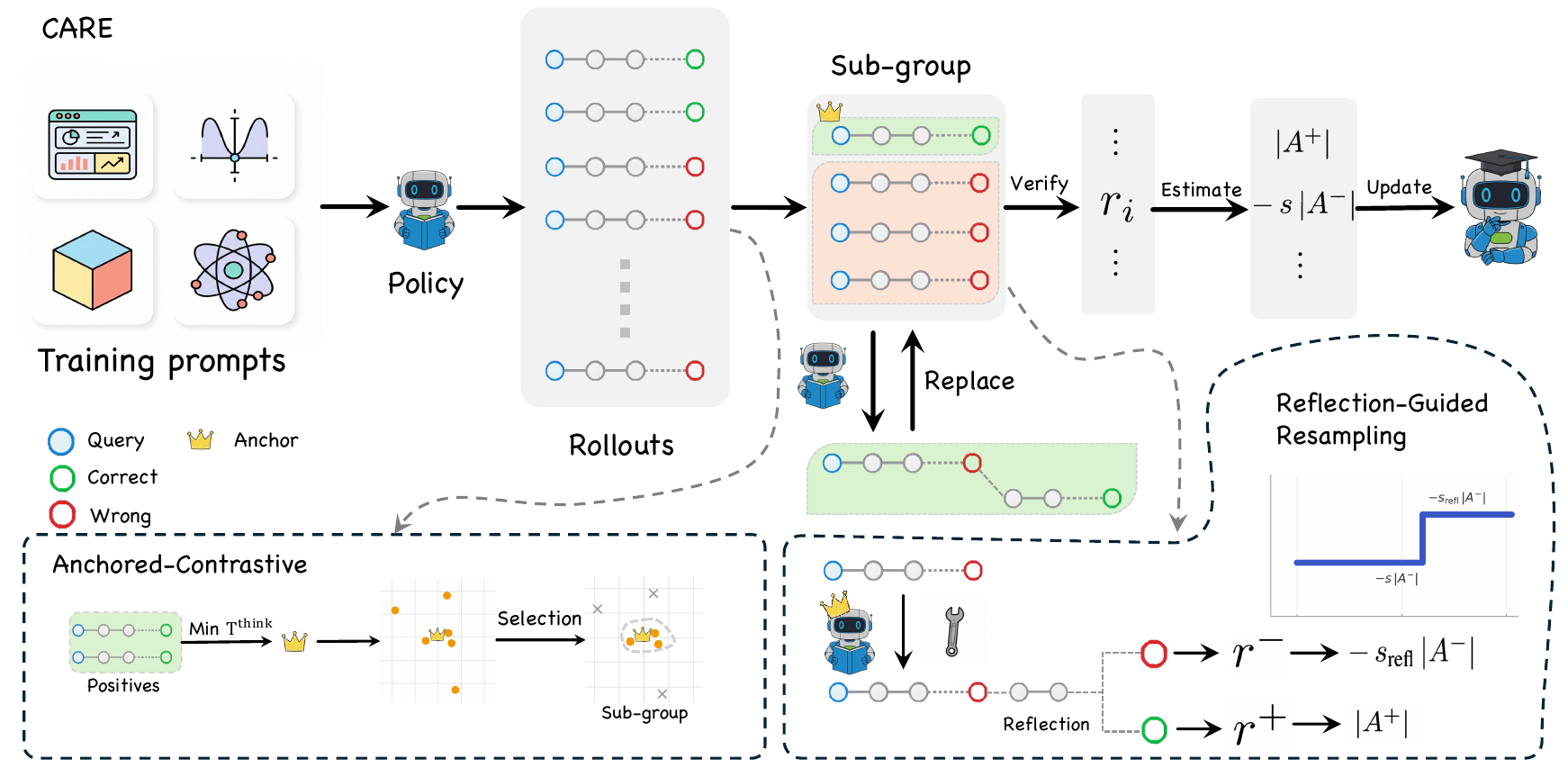}
  \vspace{-4mm}
  \captionof{figure}{\textbf{Overview of \method.} Given a multimodal prompt, the policy samples a fixed number of rollouts.
  A programmatic verifier scores the answer with rationale and supplies the reward used to update the policy.
  \emph{Anchored‑contrastive (bottom-left):} pick the anchor as the verified‑correct rollout with the shortest rationale; form a subgroup by selecting hard negatives that fail the verifier but are closest in rationale to the anchor. Advantages are normalized within this subgroup, and negatives are down‑weighted.
  \emph{Reflection‑Guided Resampling (bottom-right):} pair exactly one positive with one hard negative, insert a brief repair cue, and resample the negative once—on success replace the failure; otherwise keep it with a reduced penalty.}
  \label{fig:teaser}
\end{center}
}]

\begin{abstract}


Group-relative reinforcement learning with verifiable rewards (RLVR) often wastes the most informative data it already has—the failures. When all rollouts are wrong, gradients stall; when one happens to be correct, the update usually ignores why the others are close-but-wrong, and credit can be misassigned to spurious chains. We present \textbf{CARE} (\emph{Contrastive Anchored REflection}), a failure-centric post-training framework for multimodal reasoning that turns errors into supervision. CARE combines: (i) an \emph{anchored-contrastive objective} that forms a compact subgroup around the best rollout and a set of semantically proximate hard negatives, performs within-subgroup $z$-score normalization with \emph{negative-only} scaling, and includes an \emph{all-negative rescue} to prevent zero-signal batches; and (ii) \emph{Reflection-Guided Resampling} (RGR), a one-shot structured self-repair that rewrites a representative failure and re-scores it with the same verifier, converting near-misses into usable positives without any test-time reflection. CARE improves accuracy and training smoothness while explicitly increasing the share of learning signal that comes from failures. On Qwen2.5-VL-7B, CARE lifts macro-averaged accuracy by 4.6 points over GRPO across six verifiable visual-reasoning benchmarks; with Qwen3-VL-8B it reaches competitive or state-of-the-art results on MathVista and MMMU-Pro under an identical evaluation protocol.

\end{abstract}    
\section{Introduction}

Large Language Models (LLMs) have improved text understanding, generation, and reasoning. This progress has led to the creation of Large Multimodal Models (MLLMs) that combine vision and language. Early MLLMs, like GPT-4V \citep{openai2023gpt4v} and LLaVA \citep{liu2023visual}, show strong skills in captioning and conversational visual-question answering (VQA). However, the focus is now on more complex reasoning in areas such as mathematics, science, and engineering.
Recent studies MathVista \citep{lu2024mathvista}, MathVerse \citep{zhang2024mathverse}, MATH-Vision \citep{wang2024mathvision}, and MMMU \citep{mmmu2023} show ongoing gaps. Strong systems can reach over 70–80\% on \emph{MathVista}, but errors often occur in visually grounded, compositional reasoning. This highlights that outcome-only supervision still offers weak guidance for long-term reasoning. Reinforcement Learning with Verifiable Rewards (RLVR) presents a viable option. It uses programmatic verifiers, such as answer checkers, to give deterministic, scalable rewards \citep{mroueh2025rlvr,guo2025deepseekr1}. Group-based methods like Group Relative Policy Optimization (GRPO) \citep{shao2024deepseekmath} make RLVR workable by comparing multiple rollouts for each query \citep{shao2024deepseekmath}. However, two main issues remain substantial when budgets for rollouts are small: (1) high gradient variance and training instability, (2) flawed credit assignment. This can lead to reinforcing incorrect chains of reasoning when a correct answer is arrived at by chance.

We introduce \textbf{CARE} (\emph{Contrastive Anchored REflection}), a verifier-driven post-training framework that is explicitly failure-centric. CARE turns informative errors into learning signal in two complementary ways. First, an \emph{anchored-contrastive objective} treats the best rollout as an anchor and normalizes rewards \emph{within} a curated subgroup of hard negatives, yielding a stable, scale-aware contrast that teaches the policy to separate correct solutions from plausible failures while assigning credit in a fail-aware manner. Second, a lightweight \emph{Reflection-Guided Resampling} step performs a one-shot, error-corrective resample on a selected hard negative whenever a group contains a successful rollout, so representative mistakes are actively converted into improvements rather than passively ignored.

We validate on MathVista, MathVerse, MATH-Vision, and MMMU-Pro. Under identical inference-time budgets, \textbf{CARE} achieves state-of-the-art performance on MathVista and MMMU-Pro, and consistently surpasses strong RLVR baselines such as GRPO \cite{shao2024deepseekmath}, DAPO \citep{yu2025dapo} and GSPO \citep{zheng2025gspo}  on all benchmarks, indicating a stable and effective path to training reliable multimodal reasoners.


\noindent\textit{Contributions.} Our contributions are threefold:
\begin{itemize}
\item \textbf{Anchored Contrastive objective.} We anchor advantages to the best rollout and normalize within a subgroup of hard negatives, producing a stable, contrastive, and scale-aware learning signal with improved, failure-aware credit assignment.
\item \textbf{Reflection-Guided Resampling.} We introduce a single-shot guided resample on a selected hard negative in success-containing groups, turning representative failures into improvements rather than discarding them.
\item \textbf{Empirical gains on visual reasoning.} CARE sets new SOTA on MathVista and MMMU-Pro and delivers consistent improvements on MathVerse and MATH-Vision over established RLVR baselines.
\end{itemize}

\section{Related Works}
\label{sec:related_work}

\paragraph{Reinforcement Learning with Verifiable Reward}
RLVR fine-tunes LLMs using objective, easily checked rewards (e.g., pass/fail) instead of preference models, and has shown strong improvements on math and coding \citep{mroueh2025rlvr,shao2024deepseekmath,guo2025deepseekr1}. A common choice is Group Relative Policy Optimization (GRPO) \cite{shao2024deepseekmath}, a PPO-style method that replaces a learned value critic with group-based Monte Carlo advantages; GRPO was used for DeepSeek-R1's binary-reward training. Critic-free baselines such as RLOO \citep{ahmadian2024back} and sequence-level variants like GSPO \citep{zheng2025gspo} extend this line of work. For longer-horizon reasoning and better training stability, recent work refines GRPO: Dr.GRPO \citep{liu2025drgrpo} removes a length-inflation bias to improve token efficiency while keeping accuracy comparable; DAPO \citep{yu2025dapo} adds asymmetric clipping and selective rollouts for stability and sample efficiency; and DARS \citep{yang2025depth} allocates more rollout effort to difficult prompts with difficulty-adaptive multi-stage rollouts, improving long-tail Pass@K without increasing inference-time cost. 

\vspace{-5mm}
\paragraph{Multimodal Large Reasoning Models}
Recent efforts show that reinforcement learning fine-tuning can noticeably improve MLLMs, especially when paired with curated chain-of-thought (CoT) data and stabilizing training techniques. Bootstrapped pipelines such as MMR1 \citep{leng2025mmr1} and Vision-R1 \citep{huang2025visionr1} report gains on math-with-images benchmarks. Moving beyond outcome-only rewards, process-aware methods explicitly incentivize better reasoning chains: SophiaVL-R1 \citep{fan2025sophiavl} trains a ``thinking'' reward model with Trust-GRPO; VL-Rethinker \citep{wang2025vlrethinker} forces a self-verification ``rethinking'' step with selective replay; TWM \citep{diao2025temporal} refines query-guided temporal segments for video;  GPRO \citep{diao2026addressing} introduces gated perception-reasoning optimization; and VPPO \citep{huang2025spotlight} reweights policy updates toward visually grounded tokens. RL has also been used to equip LVLMs with task-specific skills through explicit reasoning modules and composite rewards, for example MetaSpatial \citep{pan2025metaspatial} for visual reasoning, ViGoRL \citep{vigorl2025} for spatially grounded exploration, GUI-R1 \citep{luo2025gui} for GUI agents, and R1-Track \citep{wang2025r1} for long-horizon tracking. Several works improve robustness and stability via visual perturbation during rollouts or fine-tuning \citep{ liu2025noisyrollout,li2025visionmatters, zhu2025medeyes, shen2025fine} and by sampling schemes that maintain reward variance \citep{leng2025mmr1, lian2025ui}. To reduce data requirements, two-stage pipelines leverage cold-start SFT then RL on curated reasoning corpora (e.g., Vision-R1-cold \citep{huang2025visionr1}; MMR1-SFT \citep{leng2025mmr1}). Tool- and domain-aware RL further extends multimodal reasoning to specialized settings, including visual tool use \citep{wu2025vtool}, clinically faithful radiology reports \citep{zhang2025editgrpo}, and long-horizon robotic manipulation \citep{liu2026palm, xu2025stare}, 3D scene understanding and generation \citep{deng2025best3dscenerepresentation, deng2025gaussiandwm3dgaussiandriving}.

\vspace{-4mm}
\paragraph{Multimodal Reasoning}
Prior work on multimodal reasoning often evaluates on visual-math and physics-style benchmarks such as MathVista \citep{lu2024mathvista}, MathVerse \citep{zhang2024mathverse}, MATH-Vision \citep{wang2024mathvision}, and SeePhys \citep{xiang2025seephys} which probe diagram comprehension and stepwise quantitative reasoning while providing objective answer formats (e.g., exact-match or multiple-choice) that align naturally with verifiable rewards. Beyond mathematics, MMMU-Pro \citep{yue2024mmmupro} extends assessment to college-level, multi-disciplinary problems across heterogeneous visual modalities, reducing textual shortcuts and better measuring genuine visual grounding under RLVR; we also reference the broader MMMU \citep{mmmu2023}. 
Recently, reasoning-centric corpora such as ViRL39K \citep{vl-rethinker}, LLaVA-CoT \citep{xu2025llavacot}, M3CoT \citep{chen2024m}, MMK12 \citep{mmk122025}, and R1-OneVision \cite{yang2025r1onevision} emphasize deliberate reasoning with objectively gradable end answers, which makes them a good match for RLVR training and evaluation.

\section{Methodology}
\label{sec:method}

\renewcommand{\method}{\textsc{CARE}}  

\noindent\textbf{Overview.}
We train with verifiable rewards and keep test-time decoding unchanged.
\method{} pairs an anchored-contrastive objective with a training-only reflection step that is activated only when at least one rollout succeeds.
(1) The \emph{anchored-contrastive objective} anchors sequence-level advantages to a single best rollout and normalizes them within a hard-negative subgroup. It also applies negative-penalty scaling and includes an all-negative rescue to prevent vanishing updates.
(2) \emph{Reflection-Guided Resampling} (RGR) augments training by selecting one hard negative, inserting a short repair cue into its rationale, and resampling exactly one reflected response. The policy is updated on the augmented group: reflected successes replace the original failure, while reflected failures remain negatives but receive a smaller penalty. An overview is shown in \cref{fig:teaser}.

\subsection{Problem Setting and Notation}
We study verifiable multimodal reasoning over inputs \(x=\langle \mathbf{I},\mathbf{q} \rangle\), where \(\mathbf{I}\) denotes one or more images and \(\mathbf{q}\) denotes a text query.
The policy \(\pi_\theta[y \mid x]\) emits \texttt{<think>}~\(\hat{r}\)\texttt{</think>} followed by \texttt{<answer>}~\(\hat{y}\)\texttt{</answer>}.
A programmatic verifier \(\mathcal{V}\) parses only the answer span and returns two bounded signals \(\acc[x,y]\in[0,1]\) and \(\fmt[x,y]\in[0,1]\).
We define the overall reward
\vspace{-1mm}
\begin{equation}
r_i \;=\; [1-\lambda]\cdot \acc[x,y_i] \;+\; \lambda\cdot \fmt[x,y_i], \qquad \lambda\in[0,1].
\label{eq:overall}
\end{equation}
For each prompt we draw \(G\) rollouts \(\mathcal{Y}=\{y_1,\ldots,y_G\}\) with rewards \(r_i=r[y_i]\).
Let \(T_i\) be the total number of response tokens, \(T_i^{\text{ans}}\) those inside \texttt{<answer>...</answer>}, and \(T_i^{\text{think}}\) those inside \texttt{<think>...</think>}.
We denote by \(\pi_{\text{old}}\) the frozen sampling policy for the current batch, and by \(\pi_{\text{ref}}\) a fixed reference policy used only for KL regularization.

\subsection{\method: Anchored-Contrastive Subgroup}
\label{sec:anchor-formation}
\paragraph{Anchor.}
Let \(\mathcal{P}=\{\,i:\ \acc[x,y_i]=1\,\}\) be the positive set under the verifier.
If \(\mathcal{P}\neq\varnothing\), choose the anchor

\begin{equation}
    y^+ \;=\; \arg\min_{i\in\mathcal{P}}\, T_i^{\text{think}},
\end{equation}

breaking ties toward shorter answers \(T_i^{\text{ans}}\).
If \(\mathcal{P}=\varnothing\), anchoring is bypassed and the group is handled by the all-negative rescue in \cref{sec:rescue}.

\paragraph{Hard negatives.}
Let \(\mathcal{N}\) be the hard negatives returned by the selector in \cref{sec:hybrid}.
When the number of negatives is fewer than the target subgroup size \(K\), we shrink to fit:

\begin{equation}
    \begin{aligned}
        K' \,&=\, \min\!\bigl(K,\ |\{i:\ \acc[x,y_i]=0\}|\bigr),\\
        S \,&=\, \{y^+\}\ \cup\ \{y^-_1,\ldots,y^-_{K'}\}.
    \end{aligned}
\end{equation}
\vspace{-1mm}
If \(K'=0\), we skip the update to avoid arbitrary drift under saturated rewards.

\subsection{Negative Selection}
\label{sec:hybrid}
We want negatives that are close in rationale but wrong in outcome, so the contrast targets near misses rather than mixing unrelated failure modes.
We use verifier correctness only as a binary gate, then rank failures by cosine proximity to the anchor.
Let the failure pool be \(\mathcal{F}=\{\,i:\ \acc[x,y_i]=0\,\}\).
For each \(i\in\mathcal{F}\), form a rationale embedding by mean-pooling final-layer hidden states over the \texttt{<think>} span and \(\ell_2\)-normalizing; denote the anchor's embedding by \(\tilde h_+\).
We do \emph{not} backpropagate through these embeddings (stop-gradient).
\begin{equation}
d_{\cos}(i) \;=\; 1 - \tilde h_i^\top \tilde h_+ .
\label{eq:cosdist}
\end{equation}
\vspace{-1mm}
We rank \(\mathcal{F}\) by ascending \(d_{\cos}\) and take the top \(K'\) items as hard negatives.
To reduce redundancy, we first preselect the top \(M\!>\!K'\) nearest and apply a brief farthest-first pass within this subset before returning the final \(K'\); hyperparameter details for \(M\) are provided in the Appendix.
The selected set supplies the subgroup \(S\) used by the within-group normalization in Eqs.~\eqref{eq:mu-s}--\eqref{eq:care-zscore} and the negative considered for reflection in \cref{sec:care-reflection}.

\subsection{Group-Normalized Advantages}
\label{sec:care-anchored}
\paragraph{Within-group normalization.}
Given the current subgroup \(S\), define
\begin{align}
\mu_S &= \frac{1}{|S|}\sum_{y\in S} r[y], \label{eq:mu-s}\\
\sigma_S &= \sqrt{\frac{1}{|S|}\sum_{y\in S}\bigl(r[y]-\mu_S\bigr)^2} \;+\; \varepsilon . \label{eq:sigma-s}
\end{align}
We first form \emph{raw} (pre-scaling) sequence-level advantages
\begin{equation}
A_{\mathrm{raw}}[y] \,=\, 
\begin{cases}
\displaystyle \frac{r[y]-\mu_S}{\sigma_S}, & y\in S,\\[6pt]
0, & y\notin S.
\end{cases}
\label{eq:care-zscore}
\end{equation}
A robust variant replaces \(\mu_S\) by a trimmed mean and \(\sigma_S\) by a median absolute deviation with the same \(\varepsilon\).

\paragraph{Negative-penalty scaling and update-size equalization.}
To avoid over-sharpening, we down-weight negatives and keep the anchor unchanged:
\begin{equation}
\label{eq:neg-scale}
\begin{aligned}
\forall\, y_j^- \in S:\quad A[y_j^-] &\leftarrow -\, s\,\bigl\lvert A_{\mathrm{raw}}[y_j^-]\bigr\rvert,\\
\phantom{\forall\, y_j^- \in S:\quad} A[y^+] &\leftarrow A_{\mathrm{raw}}[y^+] .
\end{aligned}
\end{equation}
with \(s\in(0,1]\).
When \(K'\!<\!K\), a global rescale \(A\leftarrow\sqrt{K/K'}\,A\) equalizes update magnitude across groups with different \(K'\) without changing direction.

\paragraph{Mechanistic signature.}
For intuition, consider a near two-level pattern within \(S\), with one anchor at \(r^+\) and negatives around \(m_-\), and let \(\Delta=r^+-m_->0\).
z-score normalization yields
\begin{equation}
A_{\mathrm{raw}}[y^+] \approx \alpha\,\sqrt{K'},
\qquad
\overline{A}_{\mathrm{raw}}[y_j^-] \approx -\,\alpha\,\frac{1}{\sqrt{K'}},
\label{eq:sqrtK-cont}
\end{equation}
with attenuation \(\alpha=\bigl(1+\tfrac{(K'+1)\operatorname{Var}_-}{\Delta^2}\bigr)^{-1/2}\in(0,1]\).
For strictly binary rewards and \(\Delta=1\) this reduces to
\begin{equation}
A_{\mathrm{raw}}[y^+] \,=\, \sqrt{K'},
\qquad
A_{\mathrm{raw}}[y_j^-] \,=\, -\,\frac{1}{\sqrt{K'}} .
\label{eq:sqrtK}
\end{equation}
\emph{Note.} The \emph{final} advantages used by the optimizer are obtained from \(A_{\mathrm{raw}}\) by applying negative-penalty scaling \(s\) (negatives only) and, when \(K'\!<\!K\), the global equalization factor \(\sqrt{K/K'}\). The proof is shown in Appendix \cref{app:proof_k}.

\subsection{Token-Weighted Policy Objective}
Let \(\rho_{i,t}=\exp\!\bigl(\log \pi_\theta[y_{i,t}\mid x]-\log \pi_{\text{old}}[y_{i,t}\mid x]\bigr)\).
Define region weights \(w_{i,t}\) as follows:
tokens inside the answer span receive weight \(1\);
tokens inside the rationale span receive weight \(\gamma^{+}\) if \(i\) is a positive sample and \(0\) if \(i\) is a negative sample.
We normalize per-token advantages as
\begin{equation}
a_{i,t} \;=\; A[y_i]\cdot \frac{w_{i,t}}{\sum_{u=1}^{T_i} w_{i,u} + \epsilon_w}.
\label{eq:token-adv}
\end{equation}
We use \(\gamma^+=0.005\) in all main results, and set \(\epsilon_w,\varepsilon>0\) as small constants for numerical stability.
The clipped surrogate is
\begin{equation}
\label{eq:ppo}
\begin{aligned}
\mathcal{L}_{\text{PG}}
= {} & -\frac{1}{|S^\star|}\sum_{i\in S^\star}\sum_{t=1}^{T_i}
\min\Bigl\{\rho_{i,t}\,a_{i,t},\\
& \operatorname{clip}\bigl(\rho_{i,t},\,1-\epsilon_{\text{low}},\,1+\epsilon_{\text{high}}\bigr)\,a_{i,t}\Bigr\},
\end{aligned}
\end{equation}
where \(S^\star=\{\,i\,:\,A[y_i]\neq 0\,\}\).
We add a regularizer
\begin{equation}
\mathcal{L} \,=\, \mathcal{L}_{\text{PG}} \;+\; \beta\,\mathrm{KL}\!\bigl[\pi_\theta \,\Vert\, \pi_{\text{ref}}\bigr].
\label{eq:total-loss}
\end{equation}

\begin{table*}[!t]
  \centering
  \caption{Evaluation of vision--language models on multimodal reasoning benchmarks. The best and second-best in each column are marked in \textbf{bold} and \underline{underlined}, respectively. A dash (\textemdash) indicates a result was not reported.}
  \label{tab:math_benchmark}
  \definecolor{ourbg}{RGB}{243,247,255}   
  \definecolor{ourblue}{RGB}{0,90,160}    
  \resizebox{0.95\textwidth}{!}{
  \begin{tabular}{l|cccccc}
    \toprule
    Models & MathVista$_{mini}$~\citep{lu2024mathvista} & MathVerse$_{mini}$~\citep{zhang2024mathverse} & MATH\mbox{-}Vision$_{full}$~\citep{wang2024mathvision} & MMMU$_{val}$~\citep{mmmu2023} & MMMU\mbox{-}Pro$_{\text{standard}}$~\citep{yue2024mmmupro} & MMMU\mbox{-}Pro$_{\text{vision}}$~\citep{yue2024mmmupro} \\
    \midrule
    \multicolumn{7}{l}{\textbf{\textit{Proprietary Models}}} \\
    GPT\mbox{-}4o~\citep{openai2024gpt4o}                     & 63.8 & 37.6 & \textemdash & \textemdash & \textemdash & \textemdash \\
    GPT\mbox{-}5\mbox{-}Nano~\citep{openai2025gpt5nano}       & 73.1 & \textemdash & 59.7 & 72.6 & \textemdash & \textemdash \\
    Claude\mbox{-}Sonnet\mbox{-}3.7~\citep{anthropic2025claude37} & 66.8 & \textemdash & 41.9 & \textbf{75.0} & \textemdash & \textemdash \\
    Gemini\mbox{-}2.0\mbox{-}Pro~\citep{google2024gemini20}   & 71.3 & 67.3 & 48.1 & 69.9 & \textemdash & \textemdash \\
    \midrule
    \multicolumn{7}{l}{\textbf{\textit{Instruct Models}}} \\
    Qwen2.5\mbox{-}VL\mbox{-}3B~\citep{bai2025qwen25vl}              & 62.0 & 47.6 & 21.2 & 46.4 & 31.1 & 21.3 \\
    Qwen2.5\mbox{-}VL\mbox{-}7B~\citep{bai2025qwen25vl}              & 68.6 & 49.2 & 22.4 & 61.3 & 36.3 & 32.8 \\
    Qwen3\mbox{-}VL\mbox{-}4B\mbox{-}Instruct~\citep{yang2025qwen3}  & 73.7 & 46.8 & 51.6 & 67.4 & \textemdash & \textemdash \\
    Qwen3\mbox{-}VL\mbox{-}8B\mbox{-}Instruct~\citep{yang2025qwen3}  & 77.2 & 62.1 & 53.9 & 69.6 & \textemdash & \textemdash \\
    InternVL3.5\mbox{-}8B\mbox{-}Instruct~\citep{wang2025internvl35} & 74.2 & \textemdash & 46.4 & 68.1 & \textemdash & \textemdash \\
    MiMo\mbox{-}VL\mbox{-}7B\mbox{-}SFT~\citep{yue2025mimovl}        & \underline{81.8} & 67.1 & 57.9 & 64.6 & 45.2 & 39.4 \\
    LLaVA\mbox{-}OneVision\mbox{-}1.5 8B~\citep{an2025llavaov15}     & 69.6 & \textemdash & 25.6 & 55.4 & 37.4 & 25.2 \\
    \midrule
    \multicolumn{7}{l}{\textbf{\textit{Reasoning Models}}} \\
    DeepEyes~\citep{zheng2025deepeyes}                    & 70.1 & 47.3 & 26.6 & \textemdash & \textemdash & \textemdash \\
    M2\mbox{-}Reasoning~\citep{wang2025m2reasoning}       & 75.0 & \textemdash & 31.5 & \textemdash & \textemdash & \textemdash \\
    Qwen3\mbox{-}VL\mbox{-}4B\mbox{-}Thinking~\citep{yang2025qwen3}  & 79.5 & \underline{75.2} & 60.0 & 70.8 & \textemdash & \textemdash \\
    Qwen3\mbox{-}VL\mbox{-}8B\mbox{-}Thinking~\citep{yang2025qwen3}  & 81.4 & \textbf{77.7} & \textbf{62.7} & \underline{74.1} & \textemdash & \textemdash \\
    MiMo\mbox{-}VL\mbox{-}7B\mbox{-}RL~\citep{yue2025mimovl}         & 81.5 & 71.5 & 60.4 & 66.7 & \underline{46.2} & \underline{40.3} \\
    Vision\mbox{-}R1\mbox{-}7B~\citep{huang2025visionr1}             & 73.5 & \textemdash & \textemdash & \textemdash & \textemdash & \textemdash \\
    Keye\mbox{-}VL\mbox{-}1.5\mbox{-}8B~\citep{keye2025vl}           & 80.7 & 59.8 & 46.0 & 71.4 & \textemdash & \textemdash \\
    InternVL3.5\mbox{-}8B\mbox{-}MPO~\citep{wang2025internvl35}      & 74.2 & \textemdash & 46.4 & 71.2 & \textemdash & \textemdash \\
    \midrule
    \multicolumn{7}{l}{\textbf{\textit{RLVR Baselines}}} \\
    Qwen2.5\mbox{-}VL\mbox{-}7B + GRPO~\citep{shao2024deepseekmath} & 68.9 & 50.8 & 25.7 & 61.1 & 36.4 & 32.8 \\
    Qwen2.5\mbox{-}VL\mbox{-}7B + DAPO~\citep{yu2025dapo}          & 72.6 & 54.2 & 29.4 & 61.6 & 37.3 & 34.7 \\
    Qwen2.5\mbox{-}VL\mbox{-}7B + GSPO~\citep{zheng2025gspo}       & 74.1 & 56.0 & 31.6 & 62.2 & 38.9 & 36.4 \\
    \midrule
    \multicolumn{7}{l}{\textbf{\textit{Our Models}}} \\
    \rowcolor{ourbg}\textbf{Qwen2.5\mbox{-}VL\mbox{-}3B {\color{ourblue}+\,\method}} & 66.5 & 49.9 & 25.0 & 47.2 & 33.1 & 24.0 \\
    \rowcolor{ourbg}\textbf{Qwen2.5\mbox{-}VL\mbox{-}7B {\color{ourblue}+\,\method}} & 74.7 & 56.8 & 32.6 & 62.5 & 39.7 & 37.1 \\
    \rowcolor{ourbg}\textbf{Qwen3\mbox{-}VL\mbox{-}8B\mbox{-}Instruct {\color{ourblue}+\,\method}} & \textbf{82.1} & 69.7 & \underline{61.7} & 71.0 & \textbf{46.7} & \textbf{41.7} \\
    \bottomrule
  \end{tabular}}
\end{table*}

\subsection{Reflection-Guided Resampling}
\label{sec:care-reflection}
RGR augments anchored-contrastive training only when a verifier-positive anchor exists.
Select exactly one positive \(y^+\) and one hard negative \(y^-\) according to \cref{sec:hybrid}.
Insert a brief repair cue inside the rationale span and resample:
\begin{quote}\small
\texttt{<think>}\\
\texttt{Your previous reasoning was incorrect. Identify the failing operation, correct it, and re-derive the necessary quantities.}\\
\texttt{Keep this section concise. Do not restate the question.}\\
\texttt{</think>}\\
\texttt{<answer> FINAL\_ANSWER\_HERE </answer>}
\end{quote}
Decode exactly one additional rollout for \(y^-\) with the same decoding hyperparameters and rescore it with the same verifier.
Run the anchored-contrastive update on the augmented group and apply two safeguards:
if the reflected sample succeeds, replace the original failed rollout in the subgroup;
if it fails, keep it as a negative and use a reduced negative-penalty scaling for that item (use \cref{eq:neg-scale} with a smaller factor \(s_{\text{refl}}\in(0,s]\)).
RGR is disabled in all-negative groups.
\emph{Default settings.} We set \(s{=}0.5\) and \(s_{\mathrm{refl}}{=}s/2\) across all main results.

\subsection{All-Negative Rescue with Partial Balance}
\label{sec:rescue}
If \(\max_i r_i\) is close to zero, the group-relative gradient can become very small.
We add a small, zero-sum pseudo-contrast on a subset \(S=\{t\}\cup\mathcal{N}\), where \(t\) is a pseudo-anchor (chosen as the failure with the highest \(\log \pi_{\text{old}}(y\!\mid\!x)\)) and \(\mathcal{N}\) contains \(K'=\min(K,|\mathcal{F}|)\) representative negatives.
On \(S\) only, assign
\begin{equation}
\label{eq:pseudoreward}
\begin{aligned}
r'[t] &= \delta,\\
r'[j] &= -\frac{\delta}{K'} \quad \text{for } j\in\mathcal{N},\\
r'[u] &= 0 \quad \text{for } u\notin S .
\end{aligned}
\end{equation}
Then compute \cref{eq:care-zscore} on \(S\) using \(r'\), apply \cref{eq:neg-scale} to negatives, and proceed with \cref{eq:token-adv,eq:ppo,eq:total-loss}.
This reproduces \cref{eq:sqrtK} and yields a well-conditioned update without modifying the true reward.
\emph{Default magnitude.} We fix the pseudo-contrast to \(\delta{=}0.1\) for all main experiments.

\section{Experiments}
\label{sec:experiments}

\begin{table*}[!t]
\centering
\caption{Ablation on the two components of \textsc{CARE}.
Anchor = anchored-contrastive objective; RGR = reflection-guided resampling.
Avg. is the macro average over the six benchmarks; $\Delta$ is the macro gain over GRPO on the same backbone.}
\label{tab:ablation_on_care}
\definecolor{ourbg}{RGB}{243,247,255}
\setlength{\tabcolsep}{4pt}
\renewcommand{\arraystretch}{1.1}
\resizebox{0.98\textwidth}{!}{
\begin{tabular}{l|cc|cccccc|cc}
\toprule
\multirow{2}{8em}{Model Config.} & \multicolumn{2}{c|}{Components}
& \multirow{2}{6em}{MathVista$_{\text{mini}}$} & \multirow{2}{6em}{MathVerse$_{\text{mini}}$} & \multirow{2}{6em}{MATH\mbox{-}Vision$_{\text{full}}$}
& \multirow{2}{4em}{MMMU$_{\text{val}}$}
& \multirow{2}{6em}{MMMU-Pro$_{\text{std}}$}
& \multirow{2}{8em}{MMMU-Pro$_{\text{vision}}$}
& \multirow{2}{4em}{\textbf{Avg.}} & \multirow{2}{6em}{\(\boldsymbol{\Delta}\) vs GRPO} \\
\cmidrule(r){2-3}
& \textbf{Anchor} & \textbf{RGR} & & & & & & & & \\
\midrule
\multicolumn{11}{l}{\textbf{\textit{Backbone: Qwen2.5-VL-3B}}} \\
\midrule
Base (Instruct)                     & \xmark & \xmark & 62.0 & 47.6 & 21.2 & 46.4 & 31.1 & 21.3 & 38.27 & -0.56 \\
GRPO (Baseline)                     & \xmark & \xmark & 63.5 & 47.9 & 22.4 & 45.4 & 31.9 & 21.9 & 38.83 & \textemdash \\
\rowcolor{ourbg} \textbf{CARE w/o RGR}
                                    & \cmark & \xmark & 65.3 & 48.2 & 24.7 & 47.4 & 32.2 & 23.1 & 40.15 & \textbf{+1.32} \\
\rowcolor{ourbg} \textbf{CARE}
                                    & \cmark & \cmark & 66.5 & 49.9 & 25.0 & 47.2 & 33.1 & 24.0 & \textbf{40.95} & \textbf{+2.12} \\
\midrule
\multicolumn{11}{l}{\textbf{\textit{Backbone: Qwen2.5-VL-7B}}} \\
\midrule
Base (Instruct)                     & \xmark & \xmark & 68.6 & 49.2 & 22.4 & 61.3 & 36.3 & 32.8 & 45.10 & -0.85 \\
GRPO (Baseline)                     & \xmark & \xmark & 68.9 & 50.8 & 25.7 & 61.1 & 36.4 & 32.8 & 45.95 & \textemdash \\
\rowcolor{ourbg} \textbf{CARE w/o RGR }
                                    & \cmark & \xmark & 74.3 & 55.4 & 31.9 & 62.8 & 38.3 & 36.4 & 49.85 & \textbf{+3.90} \\
\rowcolor{ourbg} \textbf{CARE}
                                    & \cmark & \cmark & 74.7 & 56.8 & 32.6 & 62.5 & 39.7 & 37.1 & \textbf{50.57} & \textbf{+4.62} \\
\bottomrule
\end{tabular}}
\end{table*}

\subsection{Experimental setup}
\label{sec:setup}

\textbf{Training}
We train on a visual-reasoning mixture built from ChartQA \citep{masry-etal-2022-chartqa}, Geometry3K \citep{lu-etal-2021-inter}, and ViRL39K \citep{wang2025vlrethinker}.
After deduplication and normalization, the mix contains about 49.3K unique multimodal prompts (Appendix~\Cref{app:dataset}).
Each prompt pairs an image (e.g., a chart or diagram) with a question; the model outputs a step-by-step rationale in \texttt{<think>} and the final answer in \texttt{<answer>}.

We start with a cold-start SFT stage on Vision-R1-cold \citep{huang2025visionr1} to enforce the format and bootstrap basic visual-math behavior.
RL post-training then uses our \emph{anchored-contrastive objective}: an anchor-aware hard-negative subgroup of size \(K{=}4\), z-score normalization within the subgroup, and negative-penalty scaling.
If every rollout in a subgroup fails, we add a small zero-sum pseudo-contrast so training does not freeze (\Cref{sec:rescue}).

\emph{Reflection-Guided Resampling (RGR)} is positive-paired: it only activates when a subgroup already contains at least one successful rollout.
When it triggers, we do one guided resample from a selected hard negative during training.
This is training-only; inference still uses a single decode.

Unless otherwise noted, we run five random seeds (\(n{=}5\)) and report means; 95\% confidence intervals are provided in the appendix.

\textbf{Evaluation}
We evaluate on MathVista, MathVerse, MATH-Vision, MMMU\textsubscript{val}, MMMU--Pro\textsubscript{standard}, and MMMU--Pro\textsubscript{vision}, which provide programmatically verifiable answer keys for exact-match scoring on the \texttt{<answer>} span.
All evaluations use the official LMMs-Eval \citep{zhang2024lmmsevalrealitycheckevaluation}.
All methods use a single decode per example under identical decoding hyperparameters; we do not use reflection at test time.

\textbf{Models and Baselines}
Our backbones are Qwen2.5\mbox{-}VL\mbox{-}3B, Qwen2.5\mbox{-}VL\mbox{-}7B, and Qwen3\mbox{-}VL\mbox{-}8B\mbox{-}Instruct.
On each backbone we report the Instruct model and our \method{} model.
For RLVR baselines we include GRPO \citep{shao2024deepseekmath}, DAPO \citep{yu2025dapo}, and GSPO \citep{zheng2025gspo}, trained on the same 49.3K mixture with the same rollout budget \(\mathbf{G{=}8}\), identical decoding, and a matched optimization budget (Appendix~\Cref{app:exp_details}).

\subsection{Main results}

\Cref{tab:math_benchmark} reports accuracy on six verifiable visual-reasoning benchmarks.
On MathVista\textsubscript{mini}, our best model (Qwen3\mbox{-}VL\mbox{-}8B + \method) reaches \textbf{82.1}, slightly ahead of MiMo\mbox{-}VL\mbox{-}SFT (81.8) and Qwen3\mbox{-}VL\mbox{-}8B\mbox{-}Thinking (81.4).
On MMMU-Pro\textsubscript{standard} it attains \textbf{46.7}, surpassing MiMo\mbox{-}VL\mbox{-}7B (46.2) and the other listed models.
The same model also achieves the best score on MMMU-Pro\textsubscript{vision} with \textbf{41.7}, exceeding MiMo\mbox{-}VL\mbox{-}7B (40.3) and MiMo\mbox{-}VL\mbox{-}SFT (39.4).

On MATH-Vision\textsubscript{full} and MMMU\textsubscript{val}, performance stays competitive with strong reasoning baselines.
On MathVerse\textsubscript{mini}, thinking-style decoders remain ahead, but \method{} still delivers clear gains over RLVR baselines on the same backbone.
We show representative cases in Appendix~\Cref{app:cases}.

\paragraph{Backbone scaling.}
The improvements from \method{} grow with model size.
Compared to Qwen3\mbox{-}VL\mbox{-}8B\mbox{-}Instruct, Qwen3\mbox{-}VL\mbox{-}8B + \method{} improves by +4.9 on MathVista\textsubscript{mini}, +7.6 on MathVerse\textsubscript{mini}, +7.8 on MATH-Vision\textsubscript{full}, and +1.4 on MMMU\textsubscript{val}, while also establishing strong results on both MMMU-Pro splits.

\paragraph{RLVR baselines.}
We compare \method{} against GRPO, DAPO, and GSPO on the same Qwen2.5\mbox{-}VL\mbox{-}7B backbone.
\method{} outperforms these group-relative RL baselines across datasets.
Macro-averaged over the six benchmarks, Qwen2.5\mbox{-}VL\mbox{-}7B + \method{} improves over GRPO by \textbf{+4.62} points, over DAPO by \textbf{+2.27} points, and over GSPO by \textbf{+0.70} points; the corresponding macro-average accuracies are 50.57 (\method) vs.\ 49.87 (GSPO), 48.30 (DAPO), and 45.95 (GRPO).

\begin{figure}[h]
  \centering
  \includegraphics[width=1.0\linewidth]{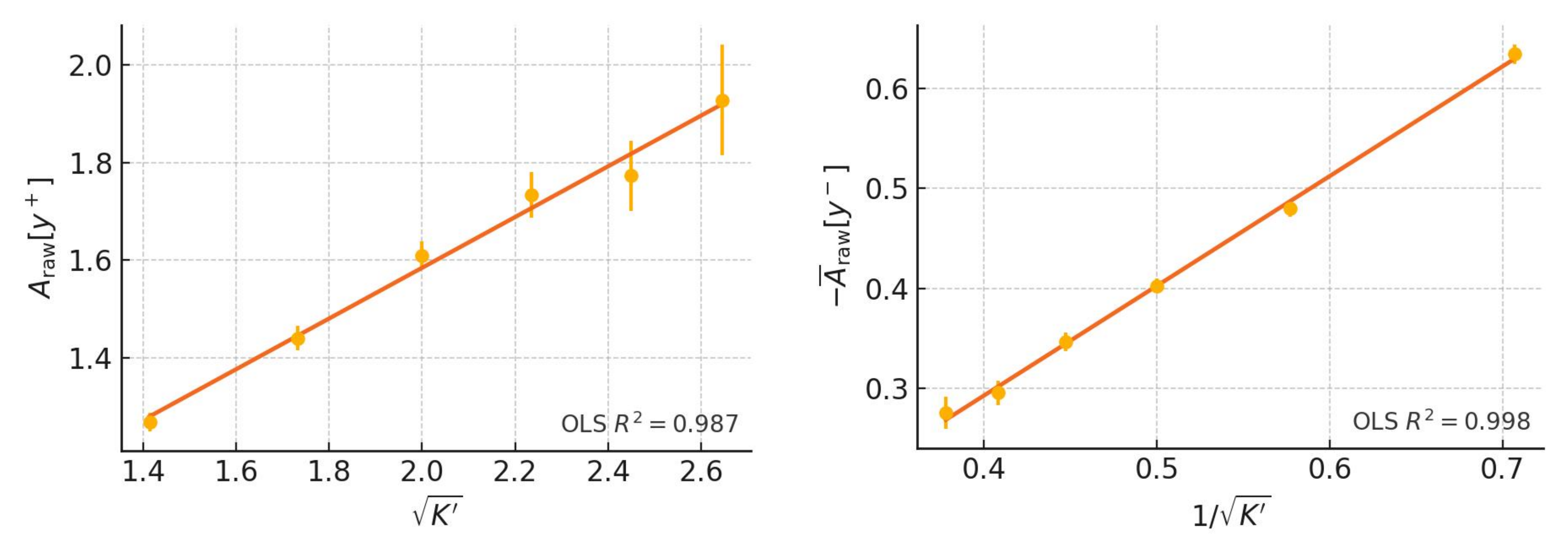}
  \caption{\textbf{$K'$ signature.}
  \emph{Left:} $A_{\mathrm{raw}}[y^+]$ vs.\ $\sqrt{K'}$.
  \emph{Right:} $-\overline{A}_{\mathrm{raw}}[y^-]$ vs.\ $1/\sqrt{K'}$.
  Markers are bucket means over groups with realized $K'\!\in\!\{2,\ldots,7\}$; vertical bars are 95\% CIs; solid lines are OLS fits on the means.}
  \label{fig:k2to7}
\end{figure}

\paragraph{Mechanistic signature.}
\method{} predicts \(A_{\mathrm{raw}}[y^+]\!\propto\!\sqrt{K'}\) and \(\overline{A}_{\mathrm{raw}}[y^-]\!\propto\!1/\sqrt{K'}\) (see \Cref{eq:sqrtK-cont,eq:sqrtK}).
\Cref{fig:k2to7} tests this by bucketing groups with realized \(K'\!\in\!\{2,\ldots,7\}\).
The bucket means follow clear linear trends consistent with the \(\sqrt{K'}\)/\(1/\sqrt{K'}\) behavior, while the error bars mainly reflect prompt-to-prompt variability.

For reference on the full sample, OLS fits for the anchor and negative sides yield \(R^2\!=\!0.223/0.555\).
The theory-normalized check \(A_{\mathrm{raw}}[y^+]/\sqrt{K'}\) vs.\ \(\widehat{\alpha}\) achieves Pearson \(r\!=\!0.998\), suggesting that most residual scatter comes from between-group variation in \(\alpha\), not a failure of the predicted scalings.

\noindent\textit{Interpretation.}
The $\sqrt{K'}$/$1/\sqrt{K'}$ law follows from z-scoring a two-level mixture with one positive and $K'$ near-miss negatives.
When the points deviate from the line, it is mostly because $\alpha$ varies across prompts, as reflected in the theory-normalized check $A_{\mathrm{raw}}[y^+]/\sqrt{K'}$ vs.\ $\widehat{\alpha}$.

\subsection{Ablation study}

\textbf{Ablation on \method{} components.}
\Cref{tab:ablation_on_care} shows that Anchor does most of the work, while RGR adds a smaller but reliable boost.
On Qwen2.5-VL-7B, the macro average rises from 45.95 (GRPO) to 49.85 with \emph{Anchor only} and to 50.57 with \emph{Anchor+RGR}.
Anchor accounts for \(84.4\%\) of the total +4.62 point gain, and RGR accounts for the remaining \(15.6\%\).
On Qwen2.5-VL-3B, the macro average climbs from 38.83 to 40.15 and then 40.95, so Anchor contributes \(62.3\%\) of the +2.12 point gain and RGR contributes \(37.7\%\).

\begin{figure}[h]
  \centering
  \includegraphics[width=\linewidth]{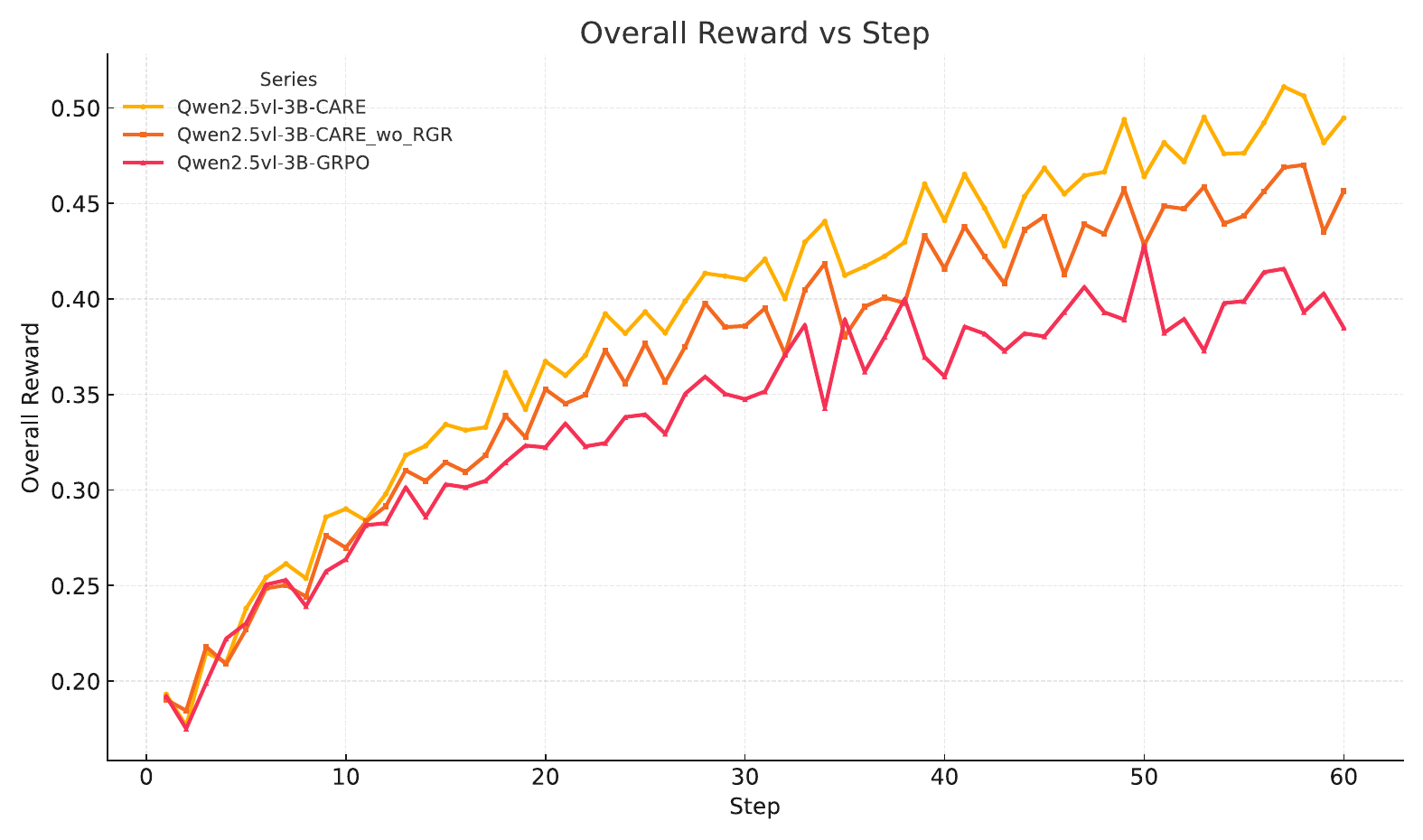}
  \caption[Qwen2.5-VL-3B training curves (Overall Reward vs Step).]{We compare \textbf{CARE}, \textbf{CARE w/o RGR}, and \textbf{GRPO} on Qwen2.5\mbox{-}VL\mbox{-}3B.
  \method{} consistently reaches higher reward at matched step budgets; \emph{CARE\_wo\_RGR} delivers most of the gain, and adding RGR yields a small, consistent, budget-neutral boost.}
  \label{fig:training_curve}
\end{figure}

The training curves in \Cref{fig:training_curve} line up with this breakdown.
Both CARE variants climb faster than GRPO, largely due to anchored within-group z-scoring with negative-only scaling.
As positives become more frequent, RGR triggers and opens a small, steady gap without extra budget.
Late in training, CARE shows fewer and shallower dips than GRPO, consistent with less volatile negative-side updates and with the all-negative \textsc{Rescue} preventing stalls.

\textbf{Ablation on reflection-guided resampling.}
We test whether RGR's gain comes from the repair cue rather than from cue-free resampling.
Conditioned on being triggered, RGR achieves a much higher success rate than two controls (\(76.6\%\) vs.\ \(19.3\%\)/\(12.8\%\)) and also yields higher macro-average accuracy (40.95 vs.\ 39.94/39.76), indicating that the cue repairs near-miss failures rather than relying on random diversity.
Inside each subgroup, anchoring (\Cref{sec:anchor-formation}) maintains a stable contrast between the best rollout and \(K'\) hard negatives.
Guided resampling targets the failing step and can turn some negatives into positives.
For reflected failures, we use a reduced negative-penalty scaling \(s_{\text{refl}}{=}s/2\) in \Cref{eq:neg-scale} to limit over-sharpening.

\begin{figure}
    \centering
    \includegraphics[width=\linewidth]{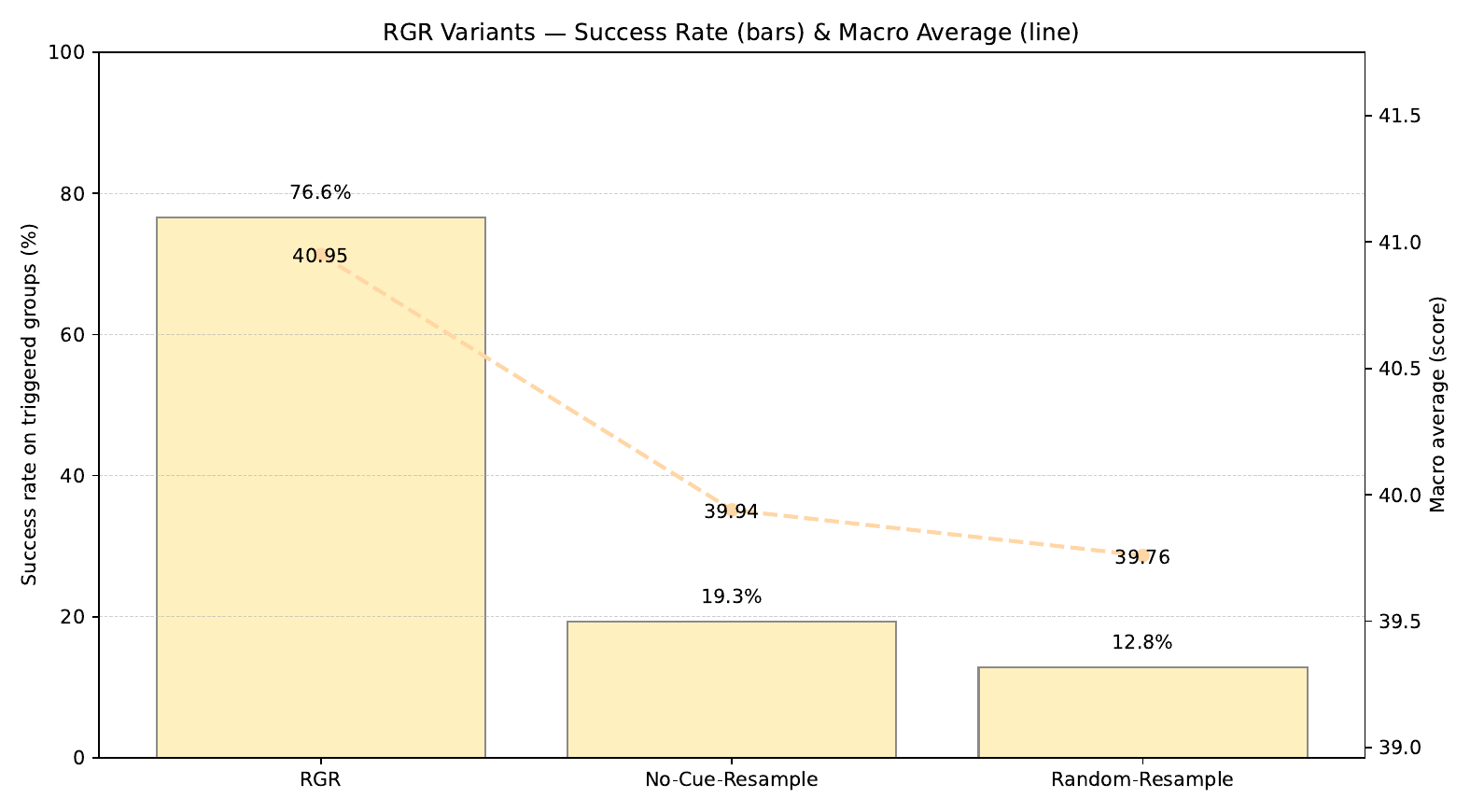}
    \caption{\textbf{Reflection success vs.\ Macro-Average (triggered-only).}
\emph{No-Cue-Resample} = resample the same hard negative \emph{without} the repair prompt;
\emph{Random-Resample} = resample \emph{without} a cue and \emph{not tied} to the original negative.
Experiments are built on Qwen2.5\mbox{-}VL\mbox{-}3B.}
    \label{fig:rgr_ablation}
\end{figure}

\begin{figure}[htbp]
  \centering
  \begin{minipage}{0.495\linewidth}
    \centering
    \includegraphics[width=\linewidth]{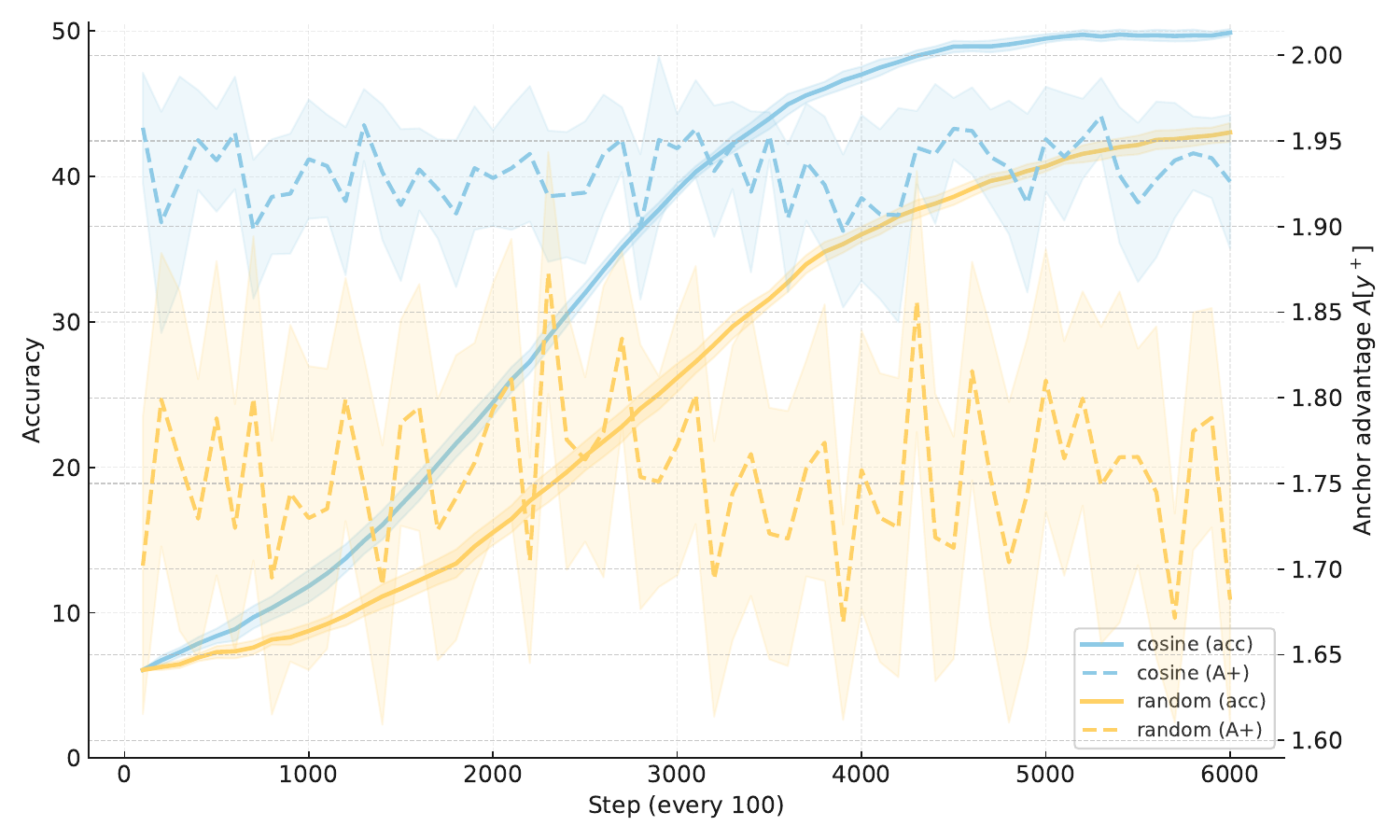}
  \end{minipage}\hfill
  \begin{minipage}{0.495\linewidth}
    \centering
    \includegraphics[width=\linewidth]{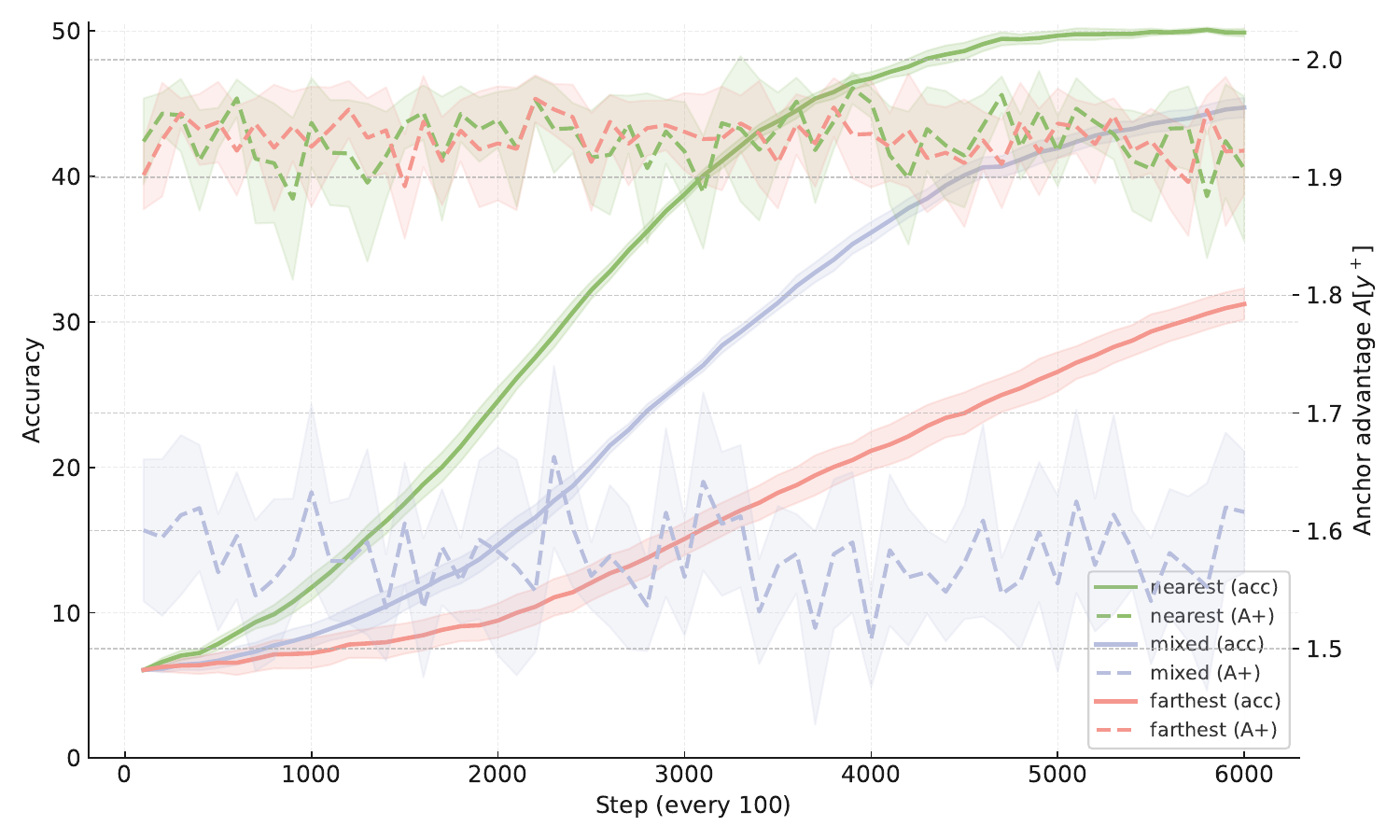}
  \end{minipage}
  \caption{\textbf{Left: Cosine vs Random.}
  Cosine-selected near-miss negatives (\textsc{cosine-top}$K'$) learn substantially faster and converge higher than \textsc{random}. \quad
  \textbf{Right: Nearest vs Mixed vs Farthest.}
  Comparing \textsc{nearest} (top-$K'$ by $d_{\cos}$), \textsc{mixed}, and \textsc{farthest}.}
  \label{fig:selection_side_by_side}
\end{figure}

\textbf{Effect of negative selection.}
We isolate the contribution of the negative selector under anchored contrast with within-group \(z\)-score normalization (Eqs.~\eqref{eq:mu-s}--\eqref{eq:care-zscore}) and cosine distance (Eq.~\eqref{eq:cosdist}).
The left \(y\)-axis reports macro-average accuracy; the right \(y\)-axis reports the anchor advantage \(A_{\mathrm{raw}}[y^+]\).
\Cref{fig:selection_side_by_side} compares \textsc{cosine-top}\(K'\) to \textsc{random}.
Starting from identical initialization, \textsc{cosine} learns faster and converges higher: it reaches about 50.57, while \textsc{random} plateaus near 43.06.
We also compare \textsc{nearest} (top-\(K'\) by \(d_{\cos}\)), \textsc{mixed} (half nearest + half farthest), and \textsc{farthest}.
The ordering is stable during training and at 6k steps---\textsc{nearest} \(>\) \textsc{mixed} \(>\) \textsc{farthest}---suggesting that mixing or selecting far negatives increases subgroup heterogeneity and weakens the contrast.
We therefore adopt \textsc{cosine-top}\(K'\) as the default selector.

\begin{figure}[htbp]
  \centering
  \begin{minipage}{0.495\linewidth}
    \centering
    \includegraphics[width=\linewidth]{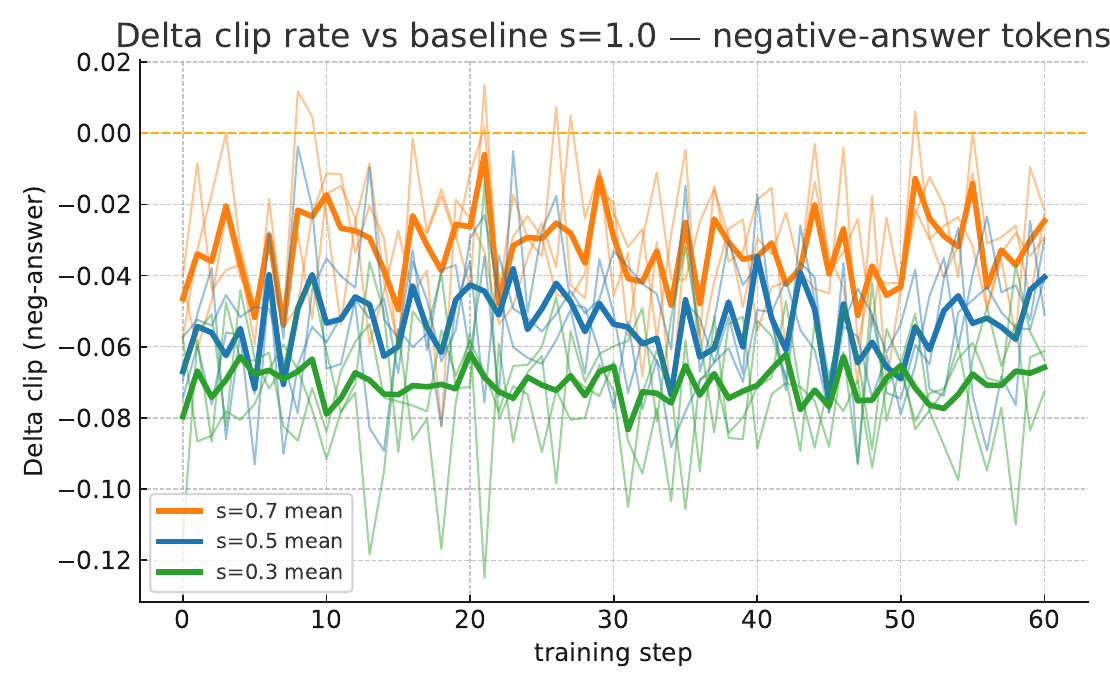}
  \end{minipage}\hfill
  \begin{minipage}{0.495\linewidth}
    \centering
    \includegraphics[width=\linewidth]{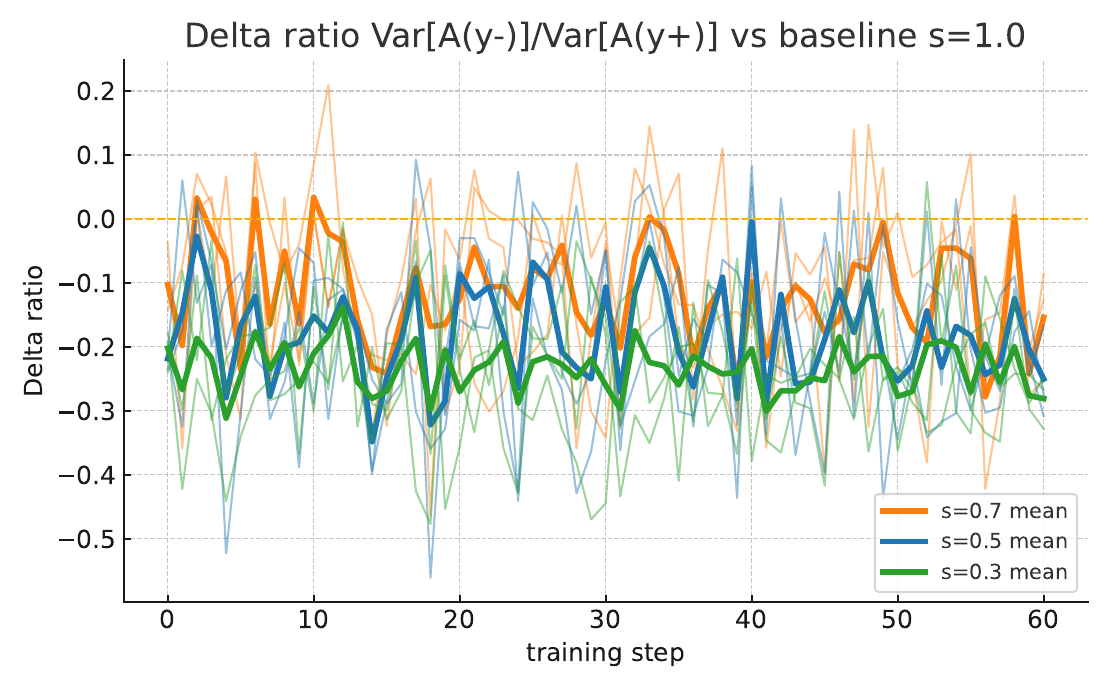}
  \end{minipage}
  \caption{Left: Stepwise \emph{delta} of the branch-clip rate on negative-answer tokens relative to the $s{=}1.0$ baseline (x-axis: training step ($\times 100$); y-axis: $\Delta$ clip). \quad
  Right: Stepwise \emph{delta} of the variance ratio $\mathrm{Var}(A[y^-])/\mathrm{Var}(A[y^+])$ relative to the $s{=}1.0$ baseline (x-axis: training step ($\times 100$); y-axis: $\Delta$ ratio).}
  \label{fig:neg_penalty}
\end{figure}

\textbf{Effect of negative-penalty scaling.}
We sweep the negative scaling factor \(s\in\{0.7,0.5,0.3\}\) against the baseline \(s{=}1.0\) under an identical budget and hyperparameters.
To increase sensitivity, we report stepwise deltas vs.\ the baseline for two diagnostics:
(i) the branch-clip rate on negative-answer tokens; and
(ii) the variance ratio \(R=\mathrm{Var}(A[y^-])/\mathrm{Var}(A[y^+])\), which tests whether scaling acts selectively on negatives.
\Cref{fig:neg_penalty} shows two consistent effects.
First, the stepwise delta in the branch-clip rate on negative-answer tokens is usually below zero and becomes more negative as \(s\) decreases, indicating fewer clipped updates where PPO is most brittle.
Second, the stepwise delta of the variance ratio \(R\) is also typically negative with the same ordering, showing that down-weighting primarily reduces negative-side variance rather than shrinking both sides equally.
Together, these results support our claim that negative-penalty scaling in \Cref{eq:neg-scale} stabilizes learning by suppressing negative-side over-updates.

\begin{figure}
    \centering
    \includegraphics[width=\linewidth]{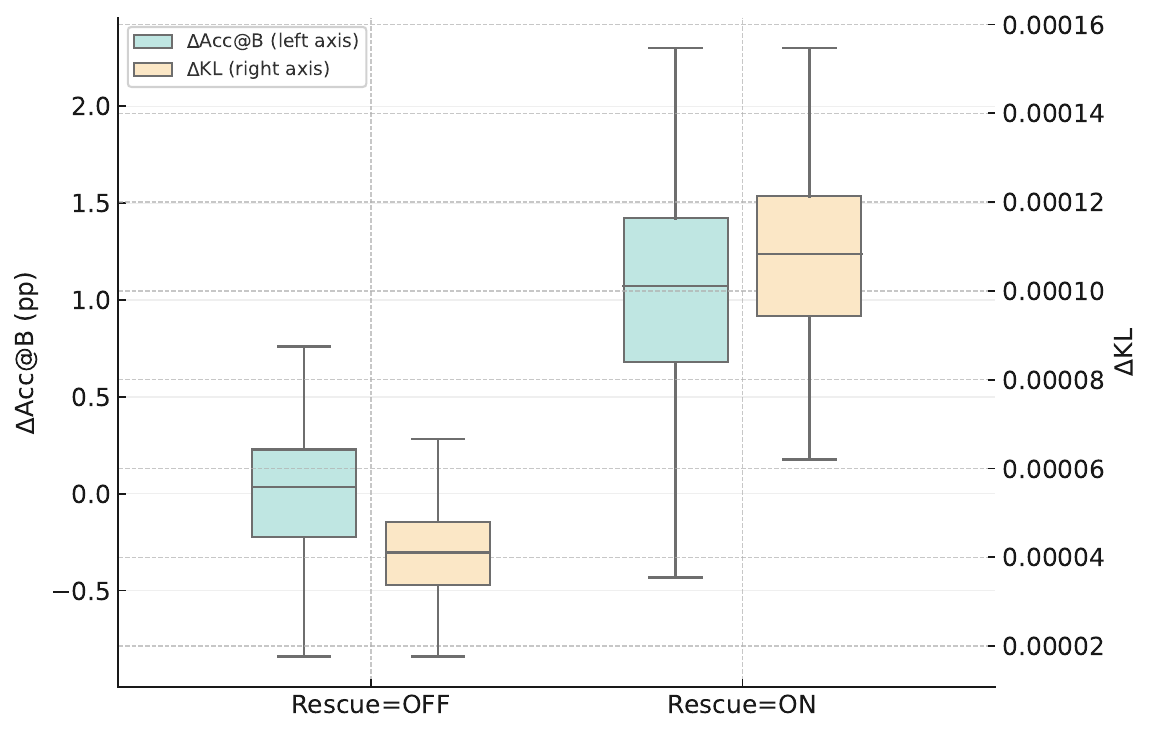}
    \caption{Dual-axis boxplot comparing \textsc{Rescue=OFF/ON} on all-negative steps under matched decoding budgets.
Left axis: $\Delta\mathrm{Acc}@B$ (percentage points); right axis: $\Delta\mathrm{KL}$.}
    \label{fig:rescue}
\end{figure}

\textbf{Effect of all-negative rescue.}
We ablate \textsc{Rescue} by toggling it ON/OFF under a shared all-negative trigger and, for each triggered step, compute the triggered-only accuracy gain over the next \(B{=}50\) steps (\(\Delta\mathrm{Acc}@B\)) together with the per-step update magnitude (\(\Delta \mathrm{KL}\)), presenting both in \Cref{fig:rescue}.
Compared with \textsc{OFF}, \textsc{Rescue=ON} shifts the \(\Delta\mathrm{Acc}@B\) distribution upward and lifts \(\Delta \mathrm{KL}\) from near-zero to small, consistent values, indicating that rescue turns stalled hard batches into short-horizon progress without increasing test-time cost.

\textbf{Effect of the region-weighted objective.}
To isolate sample efficiency, we match runs by cumulative decoded output tokens (\(T^{\text{out}}{=}T^{\text{think}}{+}T^{\text{answer}}\), prompt tokens excluded), start all curves from the same SFT checkpoint with identical decoding, and evaluate on a shared budget grid.
Resamples from RGR are counted in the same budget, and all hyperparameters are held fixed so that only region weights differ:
\emph{GRPO} (token-uniform), \emph{Answer-only} (answer \(=1\), all \texttt{<think>} \(=0\)),
\emph{Region-weighted (RW, \(\gamma^{+}{=}0.005\))} (answer \(=1\), positive \texttt{<think>} \(=\gamma^{+}\), negative/RGR-fail \texttt{<think>} \(=0\)), and
\emph{RW (fail--think unmasked)} (failing \texttt{<think>} also weighted by \(\gamma^{+}\)).

Under compute parity, RW is consistently left-shifted: for any fixed accuracy target it requires fewer tokens, achieves a larger area-under-curve, and attains a higher late-stage plateau, finishing about three points above the strongest baseline at the final budget.
The unmasked variant underperforms RW, suggesting that giving gradient credit to failing \texttt{<think>} injects noise and slows learning.
GRPO lags due to answer-credit dilution over long rationales, while Answer-only discards useful positive-rationale signal and saturates lower.

\noindent\textit{Further ablations.}
Additional controls (e.g., trigger sensitivity, selector hyperparameters \(M/K\), and decoding settings) are provided in Appendix~\Cref{app:ablation}.

\begin{figure}
    \centering
    \includegraphics[width=\linewidth]{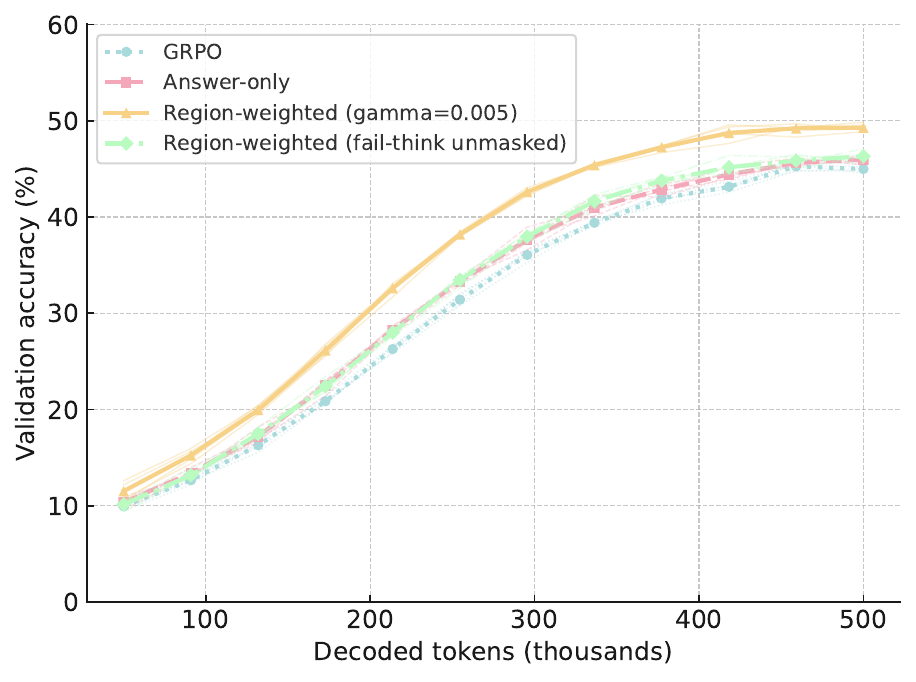}
    \caption{\textbf{Accuracy vs decoded tokens.}
All curves begin near \(6\%\) accuracy; the \emph{Region-weighted} (\(\gamma^{+}{=}0.005\)) curve reaches about \(50\%\) at the final budget, while \emph{GRPO}, \emph{Answer-only}, and \emph{Region-weighted (fail--think unmasked)} end around \(46\)--\(47\%\).
Thin lines show per-seed runs and markers show seed means; no smoothing or error bars.}
    \label{fig:rw}
\end{figure}


\section{Conclusion}

We view verifiable-reward post-training as learning from failures. Without changing the single-decode evaluation protocol, \method{} turns near-miss errors into training signal through three components. (1) An anchored, subgroup-normalized contrast objective with negative-only scaling, which pushes correct solutions away from plausible failures. (2) During training, Reflection-Guided Resampling: for success-conditioned samples, the model makes one structured repair attempt on a representative hard negative. (3) An all-negative, zero-sum rescue objective with region-weighted token credit. On MathVista, MathVerse, MATH-Vision, MMMU, and MMMU-Pro, \method{} improves accuracy. For Qwen2.5-VL-7B, it exceeds GRPO by 4.62 macro points. For Qwen3-VL-8B, it gives the strongest results in our reproducible setting on MathVista and MMMU-Pro. See Appendix~\cref{app:limitations} for limitations and future directions.


{
    \small
    \bibliographystyle{ieeenat_fullname}
    \bibliography{main}
}


\definecolor{correctgreen}{RGB}{0, 120, 0}
\definecolor{wrongred}{RGB}{180, 0, 0}
\definecolor{rgrblue}{RGB}{0, 0, 150}

\newtcolorbox{examplebox}[1]{
  colback=white,
  colframe=black,
  colbacktitle=gray!10!white,
  coltitle=black,
  title={\textbf{#1}},
  fonttitle=\bfseries\small,
  fontupper=\small,
  boxrule=0.8pt,
  arc=2mm,
  enhanced, 
  bottom=5pt,
  top=5pt
}

\clearpage
\setcounter{page}{1}
\maketitlesupplementary


\section{Notation}
\label{app:notation}

\providecommand{\acc}{\operatorname{acc}}
\providecommand{\fmt}{\operatorname{fmt}}

\begin{table}[h]
\centering
\scriptsize
\setlength{\tabcolsep}{2pt}
\renewcommand{\arraystretch}{1.03}
\begin{minipage}{0.99\columnwidth} 
\begin{tabular}{p{13mm} p{\dimexpr\linewidth-13mm\relax}} 
\hline
\textbf{Sym.} & \textbf{Meaning} \\
\hline
$x$ & Multimodal input instance. \\
$\mathbf{I}$ & Image or set of images in the input. \\
$\mathbf{q}$ & Text query in the input. \\
$\pi_\theta$ & Trainable policy that generates rationale and answer. \\
$\hat r$ & Rationale inside the \texttt{<think>} span. \\
$\hat y$ & Final answer inside the \texttt{<answer>} span. \\
$\mathcal{V}$ & Programmatic verifier of answers. \\
$\acc$ & Verifier accuracy signal. \\
$\fmt$ & Verifier format-compliance signal. \\
$r$ & Overall reward signal. \\
$r_i$ & Reward assigned to the i-th rollout. \\
$y$ & A model response. \\
$y_i$ & The i-th rollout for a given input. \\
$\mathcal{Y}$ & Collection of rollouts for a given input. \\
$G$ & Number of rollouts sampled per input. \\
$T_i$ & Total token count of the i-th response. \\
$T_i^{\text{ans}}$ & Token count inside the answer span of the i-th response. \\
$T_i^{\text{think}}$ & Token count inside the rationale span of the i-th response. \\
$\mathcal{P}$ & Indices of verifier-positive rollouts. \\
$y^+$ & Anchor rollout selected from verifier-positive rollouts. \\
$\mathcal{F}$ & Failure pool of verifier-negative rollouts. \\
$\mathcal{N}$ & Selected hard negatives. \\
$K$ & Target number of negatives in the subgroup. \\
$K'$ & Realized number of negatives in the subgroup. \\
$S$ & Anchored subgroup used for normalization and updates. \\
$S^\star$ & Indices with nonzero sequence-level advantages. \\
$M$ & Preselection size before diversity pruning among negatives. \\
$\tilde h_i$ & Normalized rationale embedding of the i-th rollout. \\
$\tilde h_+$ & Normalized rationale embedding of the anchor. \\
$d_{\cos}$ & Cosine-distance metric in the embedding space. \\
$\mu_S$ & Mean reward within the subgroup. \\
$\sigma_S$ & Standard-deviation scale within the subgroup. \\
$A[y]$ & Sequence-level advantage for a rollout. \\
$s$ & Scaling factor applied to negative advantages. \\
$s_{\mathrm{refl}}$ & Reduced scaling factor for failed reflection samples. \\
$\rho_{i,t}$ & Importance ratio at token position $t$ for the i-th rollout. \\
$w_{i,t}$ & Region weight at token position $t$ for the i-th rollout. \\
$\gamma^{+}$ & Weight for rationale tokens of verifier-positive samples. \\
$a_{i,t}$ & Per-token advantage at position $t$ for the i-th rollout. \\
$\varepsilon$ & Small constant for numerical stabilization. \\
$\epsilon_w$ & Small constant used in token-level normalization. \\
$\epsilon_{\text{low}}$ & Lower clipping threshold in the surrogate objective. \\
$\epsilon_{\text{high}}$ & Upper clipping threshold in the surrogate objective. \\
$\pi_{\text{old}}$ & Behavior policy used for importance ratios. \\
$\pi_{\text{ref}}$ & Reference policy used in KL regularization. \\
$\beta$ & Coefficient of the KL regularizer. \\
$y^-$ & Hard negative used for reflection-guided resampling. \\
\textsc{RGR} & Reflection-Guided Resampling procedure. \\
$\delta$ & Magnitude used in the all-negative rescue pseudo-contrast. \\
$t$ & Pseudo-anchor used in the all-negative rescue. \\
$r^+$ & Reward of the anchor rollout. \\
$m_-$ & Mean reward among negatives in the subgroup. \\
$\Delta$ & Reward gap between the anchor and negatives. \\
$\operatorname{Var}_-$ & Reward variance among negatives in the subgroup. \\
$\alpha$ & Attenuation factor in the mechanistic analysis. \\
\hline
\end{tabular}
\end{minipage}
\caption{\textbf{Notation.} Single-symbol entries with descriptions that avoid formulas.}
\label{tab:notation}
\end{table}

\section{Proof of $K$ Mechanistic signature}
\label{app:proof_k}

\newcommand{\E}{\mathbb{E}}
\newcommand{\Var}{\mathrm{Var}}

\theoremstyle{plain}
\newtheorem{proposition}{Proposition}
\newtheorem{corollary}{Corollary}
\theoremstyle{remark}
\newtheorem{remark}{Remark}

\paragraph{Setup.}
Let the anchored subgroup be \(S=\{y^+\}\cup\{y^-_1,\ldots,y^-_{K'}\}\). The anchor has reward \(r^+\); negatives are i.i.d.\ with mean \(m_-\) and variance \(\Var_-\). Define \(\Delta=r^+-m_->0\).
Advantages are group-normalized as in \cref{eq:mu-s,eq:sigma-s,eq:care-zscore} (we drop the tiny \(\varepsilon\) for clarity).

\begin{proposition}[Two-level $K$-signature under small dispersion]
\label{prop:ksignature}
Under the assumptions above, a plug-in (delta-method) approximation that replaces the random denominator by \(\sqrt{\E[\sigma_S^2]}\) yields
\[
\begin{aligned}
\E\,A[y^+] \;&\approx\; \alpha_{K'}\,\sqrt{K'}, 
\qquad
\E\,\overline{A}[y^-] \;\approx\; -\,\alpha_{K'}\,\frac{1}{\sqrt{K'}} ,
\end{aligned}
\]
where
\[
\alpha_{K'} 
\;=\; \frac{\Delta}{
\sqrt{\,K'\,\Var_- \;+\; \Delta^2\,}
}
\;\in\; (0,1] .
\]
\end{proposition}

\begin{proof}[Sketch]
We have \(\E[\mu_S]=m_-+\frac{\Delta}{K'+1}\) and
\[
\begin{aligned}
\E\!\bigl[(r^+-\mu_S)^2\bigr]
&= \frac{K'^2}{(K'+1)^2}\,\Delta^2 \\
&\quad+ \frac{K'}{(K'+1)^2}\,\Var_- \;.
\end{aligned}
\]
Similarly,
\[
\begin{aligned}
\E\!\bigl[(r^-_j-\mu_S)^2\bigr]
&= \Var_- \!\left( 1 - \frac{2}{K'+1}
 + \frac{K'}{(K'+1)^2} \right) \\
&\quad + \frac{\Delta^2}{(K'+1)^2}\;.
\end{aligned}
\]
Therefore
\[
\begin{aligned}
\E[\sigma_S^2]
&= \frac{1}{K'+1}\Bigl(
  \E[(r^+-\mu_S)^2]
  + K'\,\E[(r^-_j-\mu_S)^2]
\Bigr) \\
&= \frac{K'^2}{(K'+1)^2}\,\Var_- 
 \;+\; \frac{K'}{(K'+1)^2}\,\Delta^2 \;.
\end{aligned}
\]
Also \(\E[r^+ - \mu_S] = \frac{K'}{K'+1}\Delta\).
Approximating \(\E[(r^+-\mu_S)/\sigma_S] \approx \E[r^+ - \mu_S]/\sqrt{\E[\sigma_S^2]}\) gives
\[
\begin{aligned}
\E\,A[y^+] 
&\approx 
\frac{\Delta \tfrac{K'}{K'+1}}{
\sqrt{ \tfrac{K'^2}{(K'+1)^2}\,\Var_- 
     + \tfrac{K'}{(K'+1)^2}\,\Delta^2 }
} \\
&= \frac{\Delta\,\sqrt{K'}}{\sqrt{K'\Var_- + \Delta^2}}
= \alpha_{K'}\sqrt{K'} \;.
\end{aligned}
\]
For a negative \(y^-_j\), \(\E[r^-_j-\mu_S]=-\frac{\Delta}{K'+1}\), which yields the stated \(\E\,\overline{A}[y^-]\).
\end{proof}

\begin{corollary}[Binary rewards]
\label{cor:binary}
If \(r^+=1\) and all negatives satisfy \(r^-_j=0\) (i.e., \(\Var_-=0\), \(\Delta=1\)), then the $z$-scored advantages are exact:
\[
A[y^+] = \sqrt{K'}, \qquad A[y^-_j] = -\,\frac{1}{\sqrt{K'}} ,
\]
matching \cref{eq:sqrtK}.
\end{corollary}

\begin{corollary}[All-negative Rescue]
\label{cor:rescue}
Under the pseudo-reward assignment in \cref{eq:pseudoreward} on \(S=\{t\}\cup\mathcal{N}\), with \(r'[t]=\delta\), \(r'[j]=-\delta/K'\) for \(j\in\mathcal{N}\), we have \(\mu_S=0\) and
\[
\sigma_S^2=\frac{\delta^2}{K'} \;.
\]
Hence
\[
A[t]=\sqrt{K'}, \qquad A[j]=-\,\frac{1}{\sqrt{K'}} \;.
\]
Thus the rescue update reproduces the \(\sqrt{K'}\) signature exactly.
\end{corollary}

\section{Implementation Details}
\label{app:implementation}

\subsection{Datasets Details}
\label{app:dataset}

We build our visual reasoning training set by pooling three publicly available sources—ChartQA \citep{masry-etal-2022-chartqa}, Geometry3K \citep{lu-etal-2021-inter}, and ViRL39K \citep{wang2025vlrethinker}. All examples are standardized into a unified image–prompt–answer schema, followed by light cleaning and cross-corpus deduplication. The merged collection contains approximately \textbf{49.3K} unique multimodal prompts. \Cref{tab:dataset-stats} summarizes the per-source counts and shares prior to merging: \textbf{7,398} from ChartQA (15.3\%), \textbf{2,101} from Geometry3K (4.3\%), and \textbf{38,870} from ViRL39K (80.4\%), totaling \textbf{48,369} items. This mixture provides complementary coverage of chart reasoning, geometric diagram understanding, and broader visual reasoning.

\begin{table}[h]
  \centering
  \caption{Sources and sizes of the visual reasoning training set.}
  \label{tab:dataset-stats}
  \setlength{\tabcolsep}{6pt}
  \begin{tabular}{l l r r}
    \toprule
    \textbf{Dataset} & \textbf{Reference} & \textbf{Count} & \textbf{Share} \\
    \midrule
    ChartQA     & \citep{masry-etal-2022-chartqa} & \num{7398}  & 15.3\% \\
    Geometry3K  & \citep{lu-etal-2021-inter}      & \num{2101}  & 4.3\%  \\
    ViRL39K     & \citep{wang2025vlrethinker}     & \num{38870} & 80.4\% \\
    \midrule
    \textbf{Total} & & \textbf{\num{48369}} & \textbf{100.0\%} \\
    \bottomrule
  \end{tabular}
\end{table}

\begin{algorithm*}[t]
\small
\caption{\method: Anchored-Contrastive Objective with Reflection-Guided Resampling (updated)}
\label{alg:care}
\begin{algorithmic}[1]
\Require input \(x\); policy \(\pi_\theta\); old policy \(\pi_{\text{old}}\); reference \(\pi_{\text{ref}}\);
         rollout budget \(G\); target subgroup size \(K\); neg\_scale \(s\); reduced neg\_scale \(s_{\mathrm{refl}}\!\le\!s\);
         answer/rationale weight \(\gamma^{+}\); reward mix \(\lambda\in[0,1]\);
         clipping \((\epsilon_{\text{low}},\epsilon_{\text{high}})\); KL coef \(\beta\);
         normalization constants \(\varepsilon,\epsilon_w\); rescue magnitude \(\delta\)
\State Sample \(G\) rollouts \(\mathcal{Y}=\{y_i\}_{i=1}^G\); parse \texttt{<think>} and \texttt{<answer>} spans; record lengths \(T_i^{\text{think}},T_i^{\text{ans}}\)
\State For each \(y_i\): compute verifier signals \((\mathrm{acc}_i,\mathrm{fmt}_i) \leftarrow \mathcal{V}(\texttt{extract\_answer}(y_i))\); \quad set \(r_i \leftarrow (1-\lambda)\,\mathrm{acc}_i+\lambda\,\mathrm{fmt}_i\)
\State Define \(\mathcal{P}\!\leftarrow\!\{i:\mathrm{acc}_i=1\}\) (positives), \ \(\mathcal{F}\!\leftarrow\!\{i:\mathrm{acc}_i=0\}\) (failures)
\If{\(\mathcal{P}\neq\varnothing\)} \Comment{non–all–negative group}
  \State \(\displaystyle a \leftarrow y^+ \leftarrow \arg\min_{i\in\mathcal{P}} \bigl(T_i^{\text{think}},\ T_i^{\text{ans}}\bigr)\) \Comment{anchor = shortest \texttt{<think>}, tie-break shorter answer}
  \State \(K' \leftarrow \min\!\bigl(K,\ |\mathcal{F}|\bigr)\)
  \If{\(K'=0\)} \textbf{skip update} and \textbf{continue} \Comment{all-positive $\to$ avoid arbitrary drift}
  \EndIf
  \State \(\mathcal{N} \leftarrow \texttt{AnchorAwareNegatives}(\mathcal{F},\,K',\,y^+)\)
  \Comment{rank failures by cosine to anchor’s \texttt{<think>} embedding; stop-gradient; brief farthest-first de-dup}
  \State \(S \leftarrow \{y^+\}\cup\{y^-_1,\ldots,y^-_{K'}\}\);\ \ \(\tilde r(y)\leftarrow r(y)\)
  \If{reflection enabled} \Comment{Reflection-Guided Resampling (train-time only)}
    \State Choose \(y^{\mathrm{neg}\*}\in\mathcal{N}\) (e.g., hardest by \(d_{\cos}\))
    \State Insert concise repair cue into \texttt{<think>} of \(y^{\mathrm{neg}\*}\); decode once to obtain \(y^{\mathrm{rgr}}\)
    \State Parse \(y^{\mathrm{rgr}}\); compute \((\mathrm{acc}_{\mathrm{rgr}},\mathrm{fmt}_{\mathrm{rgr}})\), \(r_{\mathrm{rgr}}\leftarrow(1-\lambda)\mathrm{acc}_{\mathrm{rgr}}+\lambda\mathrm{fmt}_{\mathrm{rgr}}\)
    \If{\(\mathrm{acc}_{\mathrm{rgr}}=1\)} \Comment{success defined by verifier-positive}
      \State \(S \leftarrow \bigl(S\setminus\{y^{\mathrm{neg}\*}\}\bigr)\cup\{y^{\mathrm{rgr}}\}\); \ \(\tilde r(y^{\mathrm{rgr}})\leftarrow r_{\mathrm{rgr}}\)
    \Else \Comment{reflection fails}
      \State \(S \leftarrow S\cup\{y^{\mathrm{rgr}}\}\); \ \(\tilde r(y^{\mathrm{rgr}})\leftarrow r_{\mathrm{rgr}}\); \ mark \(y^{\mathrm{rgr}}\) for reduced scaling \(s_{\mathrm{refl}}\)
    \EndIf
  \EndIf
\Else \Comment{all–negative group}
  \State \(t \leftarrow \texttt{ProxyAnchor}(\mathcal{F}) \equiv \arg\max_{i\in\mathcal{F}} \log \pi_{\text{old}}(y_i\!\mid\!x)\)
  \State \(K' \leftarrow \min\!\bigl(K,\ |\mathcal{F}|\bigr)\);\quad \(\mathcal{N}\leftarrow \texttt{AnchorAwareNegatives}(\mathcal{F}\setminus\{t\},\,K',\,t)\)
  \State \(S \leftarrow \{t\}\cup\mathcal{N}\);\quad assign pseudo-rewards on \(S\): \(\tilde r(t)\leftarrow \delta\), \ \(\tilde r(j)\leftarrow -\delta/K'\) for \(j\in\mathcal{N}\);\quad \(a\leftarrow t\)
\EndIf
\State \(K_S \leftarrow |\{y\in S\setminus\{a\}:\ \tilde r(y)<\tilde r(a)\}|\) \Comment{negatives actually present after any reflection}
\If{\(K_S=0\)} \textbf{skip update} and \textbf{continue} \EndIf
\State \(\mu_S \leftarrow |S|^{-1}\sum_{y\in S}\tilde r(y)\);\quad \(\sigma_S \leftarrow \sqrt{|S|^{-1}\sum_{y\in S}\bigl(\tilde r(y)-\mu_S\bigr)^2}+\varepsilon\)
\State Initialize \(A(y)\leftarrow 0\) for all \(y\in\mathcal{Y}\) (and any reflected sample)
\For{each \(y\in S\)} \State \(A_{\mathrm{raw}}(y)\leftarrow\bigl(\tilde r(y)-\mu_S\bigr)/\sigma_S\) \EndFor
\State Set \(A(a)\leftarrow A_{\mathrm{raw}}(a)\)
\For{each \(y\in S\setminus\{a\}\)}
  \State \(s_y \leftarrow s\); \textbf{if} \(y\) is a failed reflected sample \textbf{then} \(s_y \leftarrow s_{\mathrm{refl}}\)
  \State \(A(y) \leftarrow -\, s_y\,\bigl|A_{\mathrm{raw}}(y)\bigr|\)
\EndFor
\State \textbf{Equalize update size:} \textbf{if} \(K_S<K\) \textbf{then} \(A(\cdot)\leftarrow \sqrt{K/K_S}\,A(\cdot)\)
\State \textbf{Region weights:} for token \(t\) in \(y_i\):
\Statex \hspace{1.2em} \(w_{i,t}=1\) if \(t\in\texttt{<answer>}\);\quad
                     \(w_{i,t}=\gamma^{+}\) if \(t\in\texttt{<think>}\) and \(\mathrm{acc}_i=1\);\quad
                     \(w_{i,t}=0\) if \(t\in\texttt{<think>}\) and \(\mathrm{acc}_i=0\)
\State Per-token advantage \(a_{i,t}\leftarrow A[y_i]\cdot \dfrac{w_{i,t}}{\sum_{u=1}^{T_i} w_{i,u}+\epsilon_w}\)
\State PPO loss with clipping:
\Statex \(\displaystyle
\mathcal{L}_{\text{PG}}
= -\tfrac{1}{|S^\star|}\sum_{i\in S^\star}\sum_{t}
\min\!\Big\{\rho_{i,t}\,a_{i,t},\,\operatorname{clip}\bigl(\rho_{i,t},\,1-\epsilon_{\text{low}},\,1+\epsilon_{\text{high}}\bigr)\,a_{i,t}\Big\}\),
where \(S^\star=\{i:A[y_i]\neq 0\}\) and \(\rho_{i,t}=\exp\!\bigl(\log\pi_\theta-\log\pi_{\text{old}}\bigr)\)
\State Add KL regularizer: \(\mathcal{L}\leftarrow \mathcal{L}_{\text{PG}}+\beta\,\mathrm{KL}\!\bigl[\pi_\theta\,\Vert\,\pi_{\text{ref}}\bigr]\); \quad update \(\theta \leftarrow \theta - \eta\,\nabla_\theta \mathcal{L}\)
\end{algorithmic}
\end{algorithm*}

\subsection{Experimental Details}
\label{app:exp_details}

\paragraph{Sampling and groups.}
Unless otherwise noted, we sample \(\mathbf{G{=}8}\) rollouts per prompt for all methods. For \method, we form an anchor‑aware hard‑negative subgroup of size \(K{=}4\) (\Cref{sec:anchor-formation,sec:hybrid}). In all main results we set
$M \;=\; 6 $. Within each subgroup we compute group‑normalized advantages (\Cref{sec:care-anchored}); negatives are down‑weighted by the factor \(s{=}0.5\) (Eq.~\eqref{eq:neg-scale}). When a verifier‑positive anchor exists, Reflection‑Guided Resampling (RGR; \Cref{sec:care-reflection}) performs one training‑only resample on the selected hard negative; reflected failures are further reduced by \(s_{\text{refl}}{=}s/2\). In all‑negative groups we enable the \emph{Rescue} mechanism with pseudo‑contrast magnitude \(\delta{=}0.1\) (\Cref{sec:rescue}). Region weights follow \Cref{eq:token-adv} with \(\gamma^{+}{=}0.005\) for positive \texttt{<think>} (note: \texttt{<reasoning>} with Qwen3VL-8B)tokens and \(0\) on negatives; answer tokens always receive weight \(1\). 

\paragraph{Optimization.}
All methods use the same PPO‐style clipped objective (\Cref{eq:ppo}) with a fixed KL regularizer to a reference policy (\Cref{eq:total-loss}). Decoding hyperparameters (e.g., temperature, nucleus/top‑$p$) are held constant across methods and reused at evaluation; inference always uses a \emph{single} decode and \emph{no} reflection.

\paragraph{Baselines and fairness.}
We re‑train GRPO~\citep{shao2024deepseekmath}, DAPO~\citep{yu2025dapo}, and GSPO~\citep{zheng2025gspo} on the same 49.3K mixture with the same rollout budget \(G{=}8\), identical decoding, and a matched optimization budget (steps, tokens, and KL schedule). Baselines keep their standard token‑uniform weighting (no region weighting) and do not use reflection. All models—ours and baselines—are trained from the same SFT cold‑start and use the same verifier signals \(\acc\) and \(\fmt\) combined by Eq.~\eqref{eq:overall}.

\paragraph{Seeds and reporting.}
Unless stated otherwise, we train with \(n{=}5\) independent random seeds and report the mean; 95\% confidence intervals are \(\bar{x}\pm 1.96\,\sigma/\sqrt{n}\) and shown in figures/tables where space permits.

\newcommand{\cliplow}{0.20}
\newcommand{\cliphigh}{0.28}
\newcommand{\klcoef}{0.02}

\begin{table*}[t]
  \centering
  \caption{Training-time configuration (shared vs.\ method-specific). “Uniform” = token-uniform weighting; “RW” = region-weighted scheme in Eq.~\eqref{eq:token-adv}.}
  \label{tab:train_config}
  \definecolor{ourbg}{RGB}{243,247,255}
  \setlength{\tabcolsep}{4pt}
  \renewcommand{\arraystretch}{1.1}
  \resizebox{0.98\textwidth}{!}{
  \begin{tabular}{l|ccccccccc}
    \toprule
    \textbf{Method} & \(\boldsymbol{G}\) & \(\boldsymbol{K}\) & Neg. scale & Rescue & Reflection (train) & Token weights & Clip\(_\text{low}\) & Clip\(_\text{high}\) & KL coef \(\boldsymbol{\beta}\) \\
    \midrule
    \rowcolor{ourbg}\textbf{CARE (ours)}      & 8 & 4 & \(s{=}0.5\) & \(\delta{=}0.1\) & Yes (1 resample) & RW (\(\gamma^{+}{=}0.005\)) & \cliplow & \cliphigh & \klcoef \\
    \rowcolor{ourbg}\textbf{CARE w/o RGR}     & 8 & 4 & \(s{=}0.5\) & \(\delta{=}0.1\) & No               & RW (\(\gamma^{+}{=}0.005\)) & \cliplow & \cliphigh & \klcoef \\
    GRPO                                       & 8 & — & —           & —                & No               & Uniform                    & \cliplow & \cliplow & \klcoef \\
    DAPO                                       & 8 & — & —           & —                & No               & Uniform                    & \cliplow & \cliphigh & - \\
    GSPO                                       & 8 & — & —           & —                & No               & Uniform                    & 3e-4 & 4e-4 & - \\
    \midrule
    \multicolumn{10}{l}{\small Shared across all methods unless noted: same data, verifier, and decoding.}\\
    \bottomrule
  \end{tabular}}
\end{table*}

\subsection{Evaluation Details}
\label{app:eval_details}

\paragraph{Common protocol.}
All evaluations use a \emph{single} decode per example under identical decoding hyperparameters for every model. We extract the text inside the \texttt{<answer>...</answer>} span and compute \emph{exact match} against the benchmark’s key using the official normalization rules (no custom post‑processing). Scores are reported per benchmark; the \emph{macro average} is the unweighted mean across all six benchmarks. The open-ended answers of benchmarks are judged by \texttt{GPT-5}.

\textbf{MathVista\textsubscript{mini}}~\citep{lu2024mathvista} targets mathematical and scientific reasoning with plots, charts, and diagrams; answers are short and programmatically verifiable. 

\textbf{MathVerse\textsubscript{mini}}~\citep{zhang2024mathverse} covers math‑intensive visual problems (algebra/geometry/physics) presented as figures or rendered text‑in‑image. 

\textbf{MATH‑Vision\textsubscript{full}}~\citep{wang2024mathvision} evaluates multi‑step math reasoning grounded in images (e.g., annotated diagrams); answers are discrete and exact‑matchable. 

\textbf{MMMU\textsubscript{val}}~\citep{mmmu2023} spans multi‑disciplinary, multi‑image questions; we report accuracy on the public validation set. 

\textbf{MMMU–Pro\textsubscript{standard}} and \textbf{MMMU–Pro\textsubscript{vision}}~\citep{yue2024mmmupro} evaluate professional‑level reasoning, with the \emph{vision} split focusing more heavily on visual content. Both provide deterministic answer keys for exact‑match evaluation.

\section{Additional Ablations}

\label{app:ablation}

\subsection{Ablation on $K$ and Equalization}
\label{sec:k_eq_ablation}

\begin{figure}
    \centering
    \includegraphics[width=\linewidth]{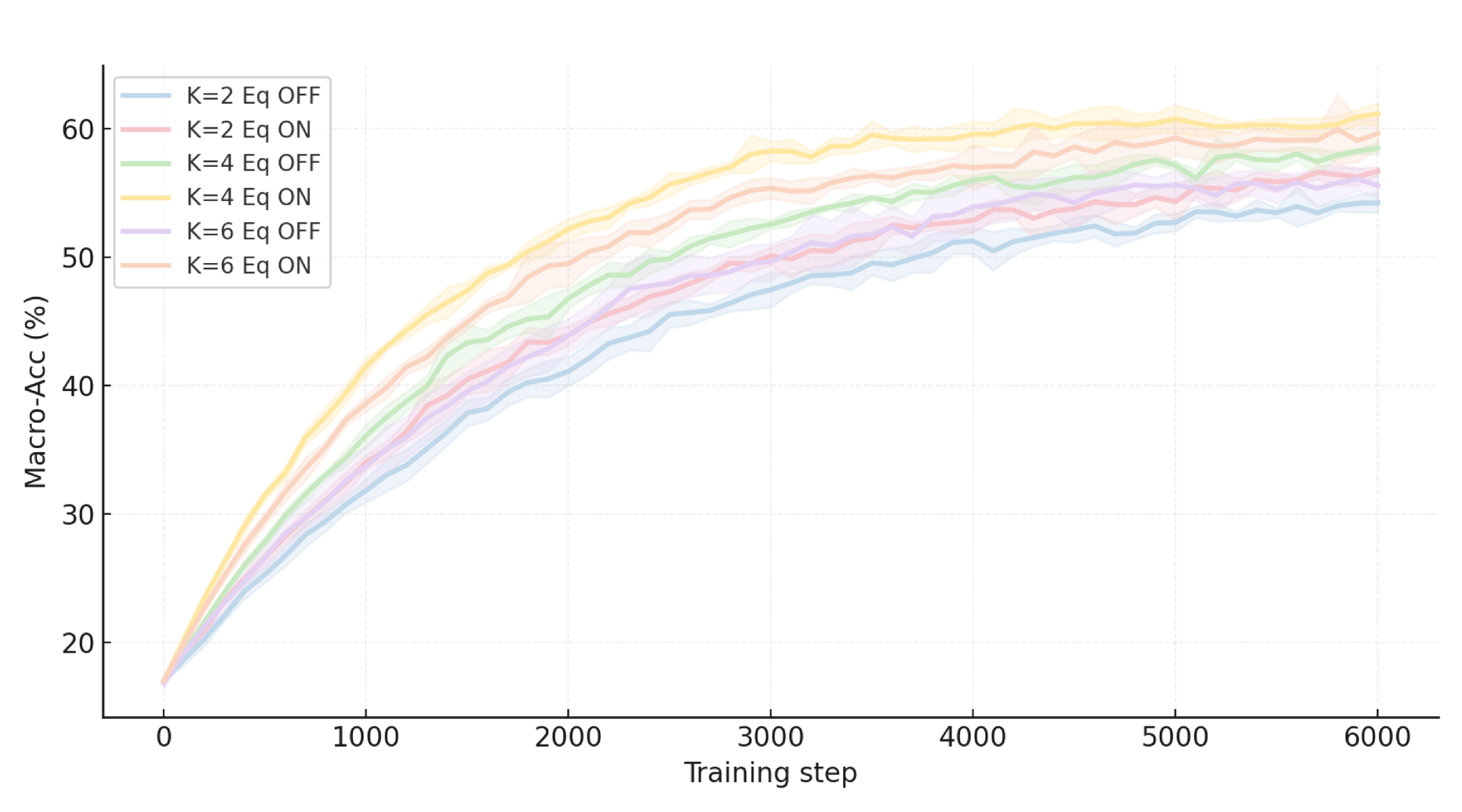}
    \caption{Val-Acc every 100 steps for $K\in\{2,4,6\}$ with/without equalization.}
    \label{fig:k_eq_acc_ci}
\end{figure}

\begin{figure}
    \centering
    \includegraphics[width=\linewidth]{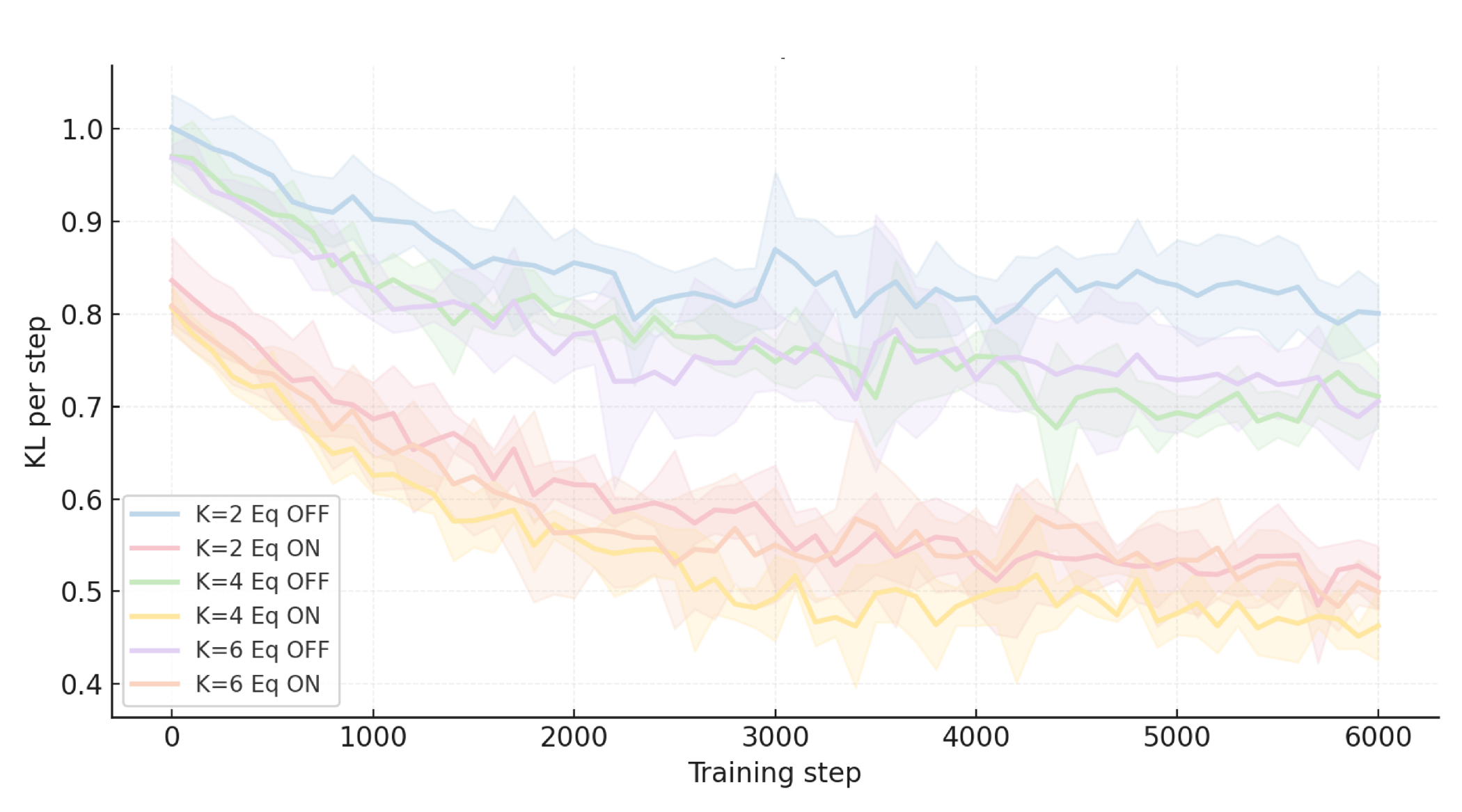}
    \caption{KL per 100 steps.}
    \label{fig:k_eq_kl_ci}
\end{figure}

We record \textbf{Val-Acc} and \textbf{KL/step} every 100 training steps. Figures~\Cref{fig:k_eq_acc_ci,fig:k_eq_kl_ci} depict the trajectories. All configurations start near \(\approx 17\%\) accuracy and differ only in the subgroup size \(K\in\{2,4,6\}\) and whether update-size equalization \(\sqrt{K/K'}\) (Eq) is enabled. Across the board, enabling Eq accelerates accuracy growth and yields higher final performance while simultaneously lowering and stabilizing KL. The strongest setting is \(K{=}4\) with Eq enabled, converging at \(\approx 61.5\%\) versus \(59.8\%\) without Eq; similar but slightly smaller gains appear for \(K{=}2\) (57.8\% vs.\ 56.8\%) and \(K{=}6\) (60.3\% vs.\ 58.5\%). The KL trajectories mirror these patterns: for \(K{=}4\), the final KL drops from \(\approx 0.70\) (Eq off) to \(\approx 0.48\) (Eq on); for \(K{=}6\), from \(\approx 0.72\) to \(\approx 0.50\); and for \(K{=}2\), from \(\approx 0.75\) to \(\approx 0.50\). We also observe narrower confidence intervals with Eq on throughout most of training, indicating reduced seed-to-seed variance and fewer regressions. These results are consistent with the mechanism that the realized hard-negative count \(K'\) often falls below \(K\): without equalization the effective update magnitude shrinks whenever \(K'\!<\!K\), slowing progress and increasing volatility, whereas the \(\sqrt{K/K'}\) factor restores a comparable update scale and produces steadier policy updates, reflected in both faster accuracy gains and smoother, lower KL. Taken together, the Acc–KL trade-off favors \(K{=}4\) with Eq on as a robust default; larger \(K\) remains competitive provided Eq is enabled, while smaller \(K\) is viable under tight budgets but saturates lower. 

\subsection{Ablation Analysis on Anchor}

\begin{figure}
    \centering
    \includegraphics[width=\linewidth]{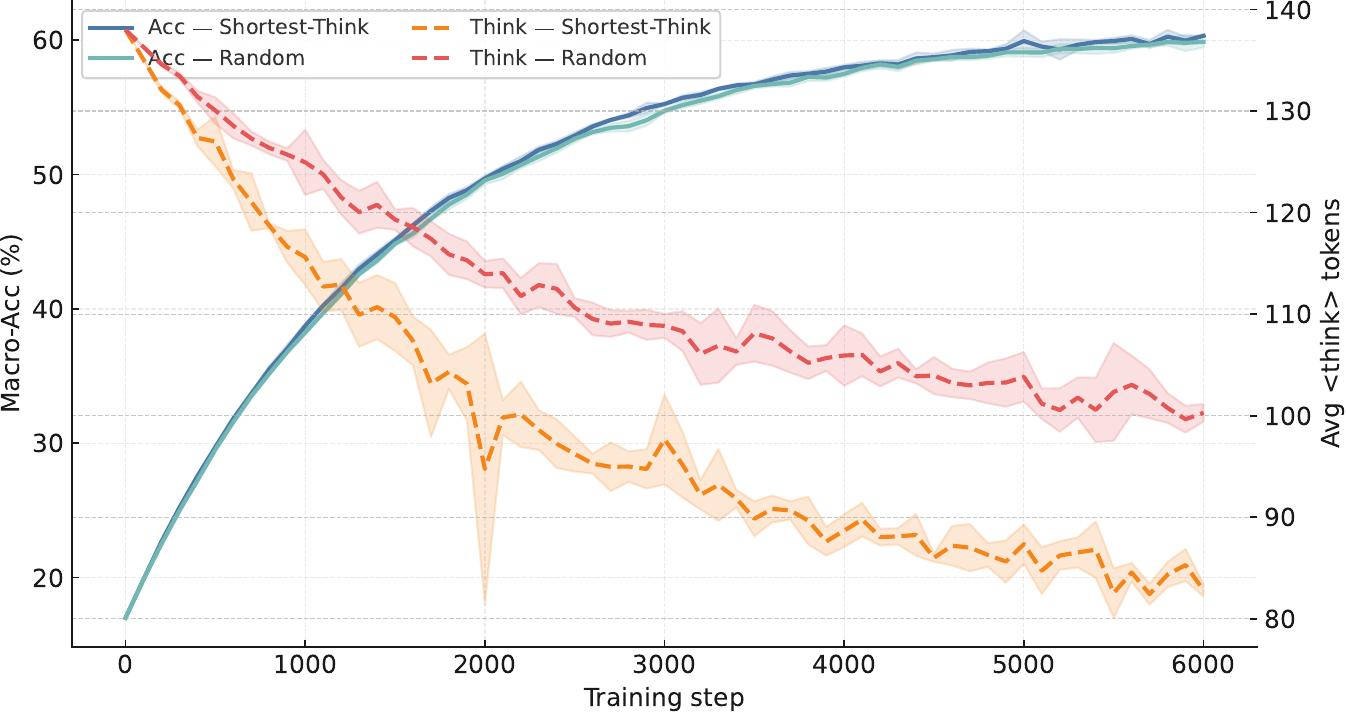}
    \caption{Val-Accuracy and Thinking token length of \emph{Shortest-Think} and \emph{Random} Anchor selection.}
    \label{fig:anchor-ablation}
\end{figure}

Holding accuracy approximately constant between the two anchor rules, we still observe a clear separation in reasoning length: with the same initial average think length at initialization, the \emph{Shortest-Think} anchor shortens the average think tokens markedly faster than a \emph{Random} anchor (\Cref{fig:anchor-ablation}). By the end of training the gap stabilizes at roughly 15–20 tokens, indicating that the model produces more compact rationales without sacrificing task accuracy. This pattern is consistent with our mechanism: choosing a concise, verifier-positive anchor concentrates the contrast on near-miss negatives and reduces credit dilution in the rationale region, yielding cleaner reasoning at comparable accuracy . Taken together, these results demonstrate that the anchoring rule controls how efficiently learning signal is allocated to the rationale; \emph{Shortest-Think} achieves shorter, more focused reasoning while maintaining the same accuracy as \emph{Random}.

\subsection{Ablation Analysis on $s$ reflection}

\begin{figure}
    \centering
    \includegraphics[width=\linewidth]{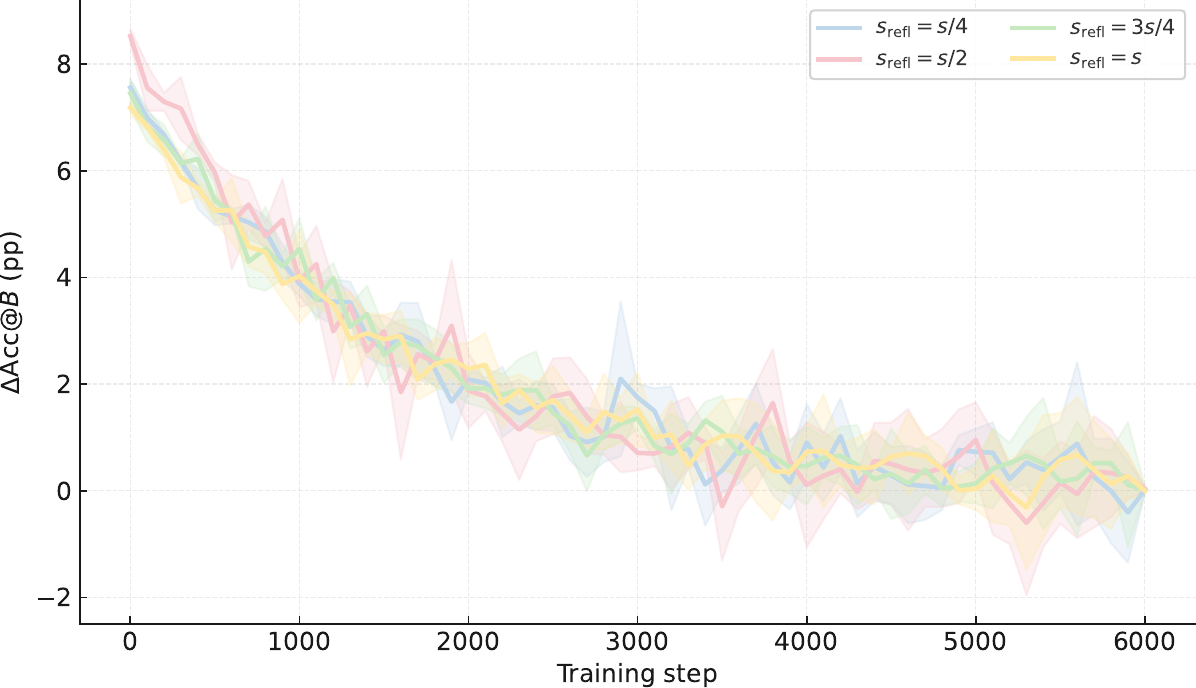}
    \caption{Across training, ACC@300 on different $s_{\mathrm{refl}}$ scale.}
    \label{fig:srefl_ablation}
\end{figure}

As shown in \Cref{fig:srefl_ablation}, across training, we observe a consistent ordering: $s_{\mathrm{refl}}{=}s/2$ yields the largest per-trigger gain, $3s/4$ is close but lower, while $s/4$ and $s$ lag behind. Early in training ($\approx$0–2k steps) the $s/2$ curve rises faster; mid-training ($\approx$2k–4k) its advantage persists with narrower confidence bands; late in training all curves flatten but $s/2$ remains on top. This pattern matches the mechanism-level intuition: too small a scaling ($s/4$) under-penalizes failed reflections and does not sufficiently move the policy off the error mode; too large a scaling ($s$) overshoots and increases brittleness; a moderate scaling ($s/2$) balances gradient magnitude and exploration radius, keeping updates within a repairable neighborhood and converting more hard errors into measurable short-horizon accuracy. We therefore adopt $s_{\mathrm{refl}}{=}s/2$ as the default choice.

\section{Future Directions and Limitations}
\label{app:limitations}

\subsection{Limitations}

While \method{} demonstrates strong robustness on verifiable reasoning tasks, its current reliance on binary correctness signals restricts its primary application to domains with objective ground truths, such as mathematics and coding, leaving the extension to open-ended generation contingent on the development of reliable reward models. Additionally, although the all-negative rescue mechanism effectively mitigates gradient collapse in sparse-reward scenarios, the framework operates most efficiently when the policy can discover at least one successful rollout to anchor the contrastive update, implying that extremely hard tasks with near-zero pass rates may still benefit from initial warm-up supervision. Finally, our heuristic preference for the shortest successful rationale is a deliberate design choice to optimize inference efficiency; while this successfully curbs verbosity, it entails a trade-off that may theoretically discourage extended reasoning chains in specific edge cases where exhaustive self-verification is strictly necessary.

\subsection{Future Directions}

\paragraph{Step‑level verification and credit.}
Augment the reward with \emph{intermediate} verifiers that check sub‑goals (e.g., variable extraction, unit conversion, geometric equivalences), enabling token‑ or span‑level advantages on \texttt{<think>} without relying on a global scalar.

\paragraph{Multi‑anchor or anchor ensembles.}
Replace the single anchor with a \emph{set} of diverse positives (e.g., by clustering successful rationales) and distribute positive credit across them.
This can mitigate brevity bias, preserve solution diversity, and improve stability on multi‑modal prompts with several correct styles.

\paragraph{Learned negative selection.}
Move beyond cosine heuristics by training a selector that predicts “fixability” or causal proximity to the anchor (e.g., via contrastive learning on success/failure edits).
Hybrid selectors could balance semantic proximity with coverage of distinct error families.

\paragraph{Adaptive subgroup sizing and scaling.}
Learn \(K'\) and the negative scaling \(s\) from data (or schedule them dynamically) to maintain a target signal‑to‑noise ratio inside the subgroup.
This could reduce sensitivity to class imbalance and reward variance across prompts.

\section{Quantitative Case Studies}
\label{app:cases}

We present detailed case studies illustrating the components of the \method{} training objective. For each example, we show:
\begin{enumerate}
    \item \textbf{The Anchor ($y^+$):} The shortest successful rollout used as the positive reference.
    \item \textbf{Hard Negatives ($y^-$):} Failed rollouts selected by cosine similarity. Note how the reasoning traces are semantically close to the anchor (e.g., correct formulas or data extraction steps) but fail due to specific hallucinations or omissions.
    \item \textbf{RGR Success:} A sample where the model, prompted with the repair cue, successfully corrects a hard negative.
\end{enumerate}

\begin{figure*}[t]
    \centering
    \begin{examplebox}{Case 1: Geometry}
        \begin{minipage}{0.35\textwidth}
            \centering
            \includegraphics[width=0.9\linewidth]{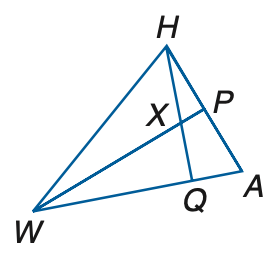}
            \vspace{1em}
            
            \textbf{Question:} If $\overline{WP}$ is a perpendicular bisector, $m \angle WHA=8 q+17$, $m \angle HWP=10+q$, $AP=6 r+4$, and $PH=22+3 r$, find $q$.
            
            \vspace{0.5em}
            \textbf{GT Answer:} 7
        \end{minipage}%
        \hfill
        \vrule width 0.5pt
        \hfill
        \begin{minipage}{0.62\textwidth}
            \textbf{\textcolor{correctgreen}{1. Anchor ($y^+$) [Positive]}} \\
            \textit{Shortest correct reasoning path.} \\
            \texttt{<think>} $\overline{WP} \perp \overline{HA} \implies \angle WPH = 90^{\circ}$. In $\Delta WHP$, angles sum to 180. Thus $(8q+17) + (10+q) + 90 = 180$. \\
            $9q + 117 = 180 \implies 9q = 63 \implies q = 7$. \texttt{</think>}\\
            \texttt{<answer>} \boxed{7} \texttt{</answer>}
            
            \vspace{0.8em}
            \hrule
            \vspace{0.8em}

            \textbf{\textcolor{wrongred}{2. Hard Negatives ($y^-$)}} \\
            \textit{Reasoning is semantically close (uses geometric properties) but hallucinates constraints.}
            
            \textbf{Neg A (Isosceles Hallucination):} \texttt{<think>} Bisector implies $\Delta WHA$ is isosceles. Base angles equal: $8q+17 = 10+q \implies 7q = -7 \implies q = -1$. \texttt{</think>} \textcolor{wrongred}{\xmark}
            
            \textbf{Neg B (Summation Error):} \texttt{<think>} Sum of given angles is 180 (ignoring right angle). $(8q+17) + (10+q) = 180 \implies 9q=153 \implies q=17$. \texttt{</think>} \textcolor{wrongred}{\xmark}

            \vspace{0.8em}
            \hrule
            \vspace{0.8em}

            \textbf{\textcolor{rgrblue}{3. Reflection-Guided Resampling (RGR)}} \\
            \textit{Corrects Neg A using the repair cue.} \\
            \texttt{<think>} ... $8q+17 = 10+q$ ... \\
            \textbf{[Correction Trigger]:} \textit{Wait, perpen. bisector means $\angle WPH=90^{\circ}$. I cannot equate the acute angles. They must sum to $90^{\circ}$.} \\
            Calculation: $(8q+17) + (10+q) = 90 \implies 9q+27=90 \implies 9q=63 \implies q=7$. \texttt{</think>} \\
            \texttt{<answer>} \boxed{7} \texttt{</answer>} \textcolor{correctgreen}{\checkmark}
        \end{minipage}
    \end{examplebox}
    \vspace{-2mm}
    \caption{Analysis of Case 1. The Anchor correctly uses the right-angle property. Hard negatives hallucinate isosceles properties or misapply the sum theorem. RGR successfully recovers from the "equality" hallucination.}
    \label{fig:case_study_geom}
\end{figure*}

\begin{figure*}[t]
    \centering
    \begin{examplebox}{Case 2: Multi-Point Data Extraction}
        \begin{minipage}{0.35\textwidth}
            \centering
            \includegraphics[width=1.0\linewidth]{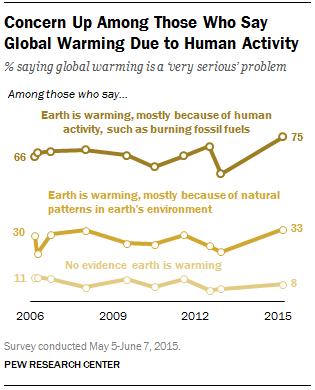}
            \vspace{1em}
            
            \textbf{Question:} What's the product of all the values in 2015?
            
            \vspace{0.5em}
            \textbf{GT Answer:} 19800
        \end{minipage}%
        \hfill
        \vrule width 0.5pt
        \hfill
        \begin{minipage}{0.62\textwidth}
            \textbf{\textcolor{correctgreen}{1. Anchor ($y^+$) [Positive]}} \\
            \texttt{<think>} Locating 2015. Three lines visible. \\
            1. Top (Human activity): 75. \\
            2. Middle (Natural patterns): 33. \\
            3. Bottom (No evidence): 8. \\
            Product: $75 \times 33 \times 8 = 2475 \times 8 = 19800$. \texttt{</think>}\\
            \texttt{<answer>} \boxed{19800} \texttt{</answer>}
            
            \vspace{0.8em}
            \hrule
            \vspace{0.8em}

            \textbf{\textcolor{wrongred}{2. Hard Negatives ($y^-$)}} \\
            \textit{Correct formula application but fails visual grounding.}
            
            \textbf{Neg A (Visual Omission):} \texttt{<think>} Reading 2015. Top is 75. Middle is 33. No other lines clearly visible. Product: $75 \times 33 = 2475$. \texttt{</think>} \textcolor{wrongred}{\xmark}
            
            \textbf{Neg B (Data Hallucination):} \texttt{<think>} Values are 75, 33, and the bottom looks like 10. Product: $75 \times 33 \times 10 = 24750$. \texttt{</think>} \textcolor{wrongred}{\xmark}

            \vspace{0.8em}
            \hrule
            \vspace{0.8em}

            \textbf{\textcolor{rgrblue}{3. Reflection-Guided Resampling (RGR)}} \\
            \textit{Corrects Neg A (Omission).} \\
            \texttt{<think>} ... Product is 2475 ... \\
            \textbf{[Correction Trigger]:} \textit{Re-checking the chart. I missed the bottom beige line labeled 'No evidence'. The endpoint at 2015 is explicitly 8.} \\
            Revised values: 75, 33, 8. Calculation: $2475 \times 8 = 19800$. \texttt{</think>} \\
            \texttt{<answer>} \boxed{19800} \texttt{</answer>} \textcolor{correctgreen}{\checkmark}
        \end{minipage}
    \end{examplebox}
    \vspace{-2mm}
    \caption{Analysis of Case 2. Hard negatives often miss the bottom trend line or misread the axis. RGR triggers a visual re-verification step ("Re-checking the chart") to recover the missing data point.}
    \label{fig:case_study_chart}
\end{figure*}

\begin{figure*}[t]
    \centering
    \begin{examplebox}{Case 3: Geometry (Circle Properties)}
        \begin{minipage}{0.35\textwidth}
            \centering
            \includegraphics[width=0.8\linewidth]{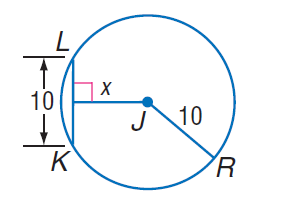}
            \vspace{1em}
            
            \textbf{Question:} Find $x$.
            
            \vspace{0.5em}
            \textbf{GT Answer:} $5\sqrt{3}$ (or approx 8.66)
        \end{minipage}%
        \hfill
        \vrule width 0.5pt
        \hfill
        \begin{minipage}{0.62\textwidth}
            \textbf{\textcolor{correctgreen}{1. Anchor ($y^+$) [Positive]}} \\
            \textit{Correctly identifies radius and applies Pythagorean theorem.} \\
            \texttt{<think>} The radius is $JR=10$. The segment from center $J$ to any point on the circle is 10, so hypotenuse to $L$ is 10. The perpendicular segment $x$ bisects chord $LK$. Total chord is 10, so leg is 5. \\
            Pythagoras: $x^2 + 5^2 = 10^2 \implies x^2 + 25 = 100 \implies x^2 = 75$. \\
            $x = \sqrt{75} = 5\sqrt{3}$. \texttt{</think>}\\
            \texttt{<answer>} \boxed{5\sqrt{3}} \texttt{</answer>}
            
            \vspace{0.8em}
            \hrule
            \vspace{0.8em}

            \textbf{\textcolor{wrongred}{2. Hard Negatives ($y^-$)}} \\
            \textit{Valid steps (identifying the triangle) but fails on geometric rules.}
            
            \textbf{Neg A (Equilateral Hallucination):} \texttt{<think>} Chord is 10, Radius is 10. The triangle formed by the chord and center is equilateral. The line $x$ splits the side. Therefore $x$ is simply half of the side length. $x = 10 / 2 = 5$. \texttt{</think>} \textcolor{wrongred}{\xmark}
            
            \textbf{Neg B (Hypotenuse Confusion):} \texttt{<think>} We have a right triangle. The legs are the distance $x$ and the chord length 10. The radius 10 is the hypotenuse. Wait, actually, we form a triangle with sides 10 and 5. $x = \sqrt{10^2 + 5^2} = \sqrt{125} = 5\sqrt{5}$. \texttt{</think>} \textcolor{wrongred}{\xmark}

            \vspace{0.8em}
            \hrule
            \vspace{0.8em}

            \textbf{\textcolor{rgrblue}{3. Reflection-Guided Resampling (RGR)}} \\
            \textit{Corrects Neg A (Equilateral assumption).} \\
            \texttt{<think>} ... Triangle is equilateral, so $x$ is half the side, which is 5. ... \\
            \textbf{[Correction Trigger]:} \textit{Check the altitude calculation. In an equilateral triangle, the altitude is not half the side; it relates via $\sqrt{3}$.} \\
            Re-calculating: This is a right triangle with hypotenuse 10 (radius) and base 5 (half-chord). \\
            $x^2 = 100 - 25 = 75$. $x = \sqrt{75} = 5\sqrt{3}$. \texttt{</think>} \\
            \texttt{<answer>} \boxed{5\sqrt{3}} \texttt{</answer>} \textcolor{correctgreen}{\checkmark}
        \end{minipage}
    \end{examplebox}
    \vspace{-2mm}
    \caption{Analysis of Case 3. The model often mistakenly assumes $x$ is a median of length 5 (Neg A) or adds squares instead of subtracting (Neg B). RGR forces a check on the specific right-triangle relationship.}
    \label{fig:case_study_circle}
\end{figure*}

\begin{figure*}[t]
    \centering
    \begin{examplebox}{Case 4: Chart Legend Association}
        \begin{minipage}{0.35\textwidth}
            \centering
            \includegraphics[width=0.8\linewidth]{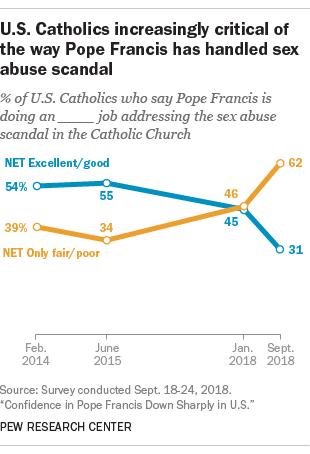}
            \vspace{1em}
            
            \textbf{Question:} What's the blue graph represent?
            
            \vspace{0.5em}
            \textbf{GT Answer:} NET Excellent/good
        \end{minipage}%
        \hfill
        \vrule width 0.5pt
        \hfill
        \begin{minipage}{0.62\textwidth}
            \textbf{\textcolor{correctgreen}{1. Anchor ($y^+$) [Positive]}} \\
            \texttt{<think>} I need to identify the label for the blue line. \\
            1. The blue line starts at 54\% and ends at 31\%. \\
            2. The text colored in blue/teal directly above the start of the line says "NET Excellent/good". \\
            3. The question asks for the representation. \\
            Result: It represents those saying Pope Francis is doing an excellent or good job. \texttt{</think>}\\
            \texttt{<answer>} \boxed{\text{NET Excellent/good}} \texttt{</answer>}
            
            \vspace{0.8em}
            \hrule
            \vspace{0.8em}

            \textbf{\textcolor{wrongred}{2. Hard Negatives ($y^-$)}} \\
            \textit{Correct visual attention (looking at lines) but semantic mismatch.}
            
            \textbf{Neg A (Trend Description):} \texttt{<think>} The blue line goes down significantly from 54\% to 31\%. The title says Catholics are "increasingly critical". Therefore, the blue line represents the "Critical" view or the "Sex abuse scandal". \texttt{</think>} \textcolor{wrongred}{\xmark}
            
            \textbf{Neg B (Label Swap):} \texttt{<think>} There are two lines. The blue one is the lower one at the end (2018). The label nearest to the end of the graph is "NET Only fair/poor" (orange). I will associate the blue line with fair/poor performance. \texttt{</think>} \textcolor{wrongred}{\xmark}

            \vspace{0.8em}
            \hrule
            \vspace{0.8em}

            \textbf{\textcolor{rgrblue}{3. Reflection-Guided Resampling (RGR)}} \\
            \textit{Corrects Neg A (Trend vs Label).} \\
            \texttt{<think>} ... The blue line shows the critical trend mentioned in the title. ... \\
            \textbf{[Correction Trigger]:} \textit{Distinguish between the chart title and the specific legend entry. Look for text matching the line's color.} \\
            Correction: The title describes the trend, but the legend defines the data. The text "NET Excellent/good" is colored teal/blue, matching the line. The line represents positive sentiment. \texttt{</think>} \\
            \texttt{<answer>} \boxed{\text{NET Excellent/good}} \texttt{</answer>} \textcolor{correctgreen}{\checkmark}
        \end{minipage}
    \end{examplebox}
    \vspace{-2mm}
    \caption{Analysis of Case 4. Hard negatives frequently conflate the chart's narrative title with the specific data series (Neg A) or swap legends due to proximity (Neg B). RGR steers the model to verify color-text alignment.}
    \label{fig:case_study_pope}
\end{figure*}

\end{document}